\documentclass[11pt]{article}

\usepackage[preprint]{acl}

\usepackage{times}
\usepackage{latexsym}

\usepackage[T1]{fontenc}

\usepackage[utf8]{inputenc}

\usepackage{microtype}

\usepackage{inconsolata}

\usepackage{graphicx}
\usepackage{array}
\usepackage{booktabs}
\usepackage{amssymb}

\title{I would rather quit NLP than read another paper like this:\\The rise of antithesis in NLP papers}

\author{Olga Zamaraeva \quad Adri\'{a}n Gude\thanks{\ \ Equal contribution.} \quad Roi Santos-R\'{\i}os\footnotemark[1] \quad Carlos G\'{o}mez-Rodr\'{\i}guez \\
  Universidade da Coru\~{n}a, CITIC \\
  \texttt{\{olga.zamaraeva, adrian.lopez.gude, roi.santos.rios, carlos.gomez\}@udc.es}}

\begin{document}
\maketitle
\begin{abstract}
For better or worse, LLMs are by now used routinely for scientific writing.\footnote{This paper is no exception; we did use AI to assist with writing some of the sections (see Acknowledgments).} Many have noticed that recent models fill papers with unnecessary antithesis, stating over and over what the work does not do, in ways that do not contribute to its precision or quality of expression and annoy reviewers \emph{rather than impressing them}. We study the construction \emph{rather than} in ACL papers from 2019, ACL-style arXiv papers from 2026, and papers written by GPT models from the same titles and abstracts. Its rate in 2026 is seven times the 2019 rate, and higher still in the GPT papers. Two annotators, blind to the source, find almost no 2019 use \emph{annoying} and about one in ten 2026 uses; they seldom agree on which, yet about half of 2026 papers contain a use that annoys each of them. \emph{Annoying} uses present the rejected alternative less favorably than legitimate uses. Raters of preference data and open reward models favor the construction, and an instruction to be honest promotes it. We conjecture that it is a side effect of post-training on pairwise preferences, which credit a disavowal in a single response and cannot register its cost across a text.
\end{abstract}

\section{Introduction}
\label{sec:intro}

Large language models are increasingly part of the production process for scientific text, and their use is leaving measurable traces in academic writing, from the frequency of stylistic vocabulary to lexical choice and sentiment \citep{kobak2025delving,geng-trotta-2025-human,tudino-qin-2024-corpus,mak-walasek-2025-style}. 

One recent pattern is the repeated use of contrastive or corrective formulations that define a claim partly by stating what it is \emph{not}. The website \emph{AI tells} \citep{druck-paredes-smith-2026-ai-tells}\footnote{https://graphite.io/five-percent/research/ai-tells} calls this tendency \emph{definition by contrast} and reports that recent LLMs produce such corrective framing, including constructions of the form \emph{X rather than Y}, more than before. At the same time, the authors and some of their colleagues have noticed being \emph{annoyed} by seeing the construction in the NLP papers they were reviewing in the last ARR cycle; some reviews called it out explicitly and complained that it is a sign of a degraded quality of the papers. Program chairs of NLP venues now discuss ``AI-trope-ridden'' writing in submissions, and submissions generated wholesale \citep{durrett2026colm}, and readers outside research name contrastive constructions, \emph{rather than} among them, as marks of LLM writing that they find annoying \citep{wikipedia2026signs}. \citet{gonzales2026slanguage} furthermore discusses possible information processing-related harms associated with overusing negation constructions.

\emph{Rather than} is an ordinary English construction with established use for expressing alternatives and preferences \citep{thompson1972instead,silvennoinen2023not}. Beyond how often LLMs use \emph{rather than} and its kin, the question is how this increased use came about, why it can be \emph{annoying}, and how it affects the readers and their perception of the construction itself, the scientific prose and ultimately, their trust in the field.

Answering all of these questions should be a subject of a larger study involving psychologists as well as specialists privy to the details of post-training of commercial systems such as Claude and GPT. Here, we attempt to address some of the preliminaries, in four questions. \textbf{Q1}: Has \emph{rather than} become more frequent in NLP papers, and do some\footnote{We hypothesize from the start that this will be true only with respect to some commercial models due to the specific post-training they may be using.} LLMs use it more often than human authors? \textbf{Q2}: Do readers find some of its uses \emph{annoying}, and how likely is a reader to meet an \emph{annoying} use in a paper? \textbf{Q3}: What distinguishes \emph{annoying} uses from legitimate ones? \textbf{Q4}: Where does the construction come from: does the preference data used in post-training reward it?

We compare ACL papers from 2019, before widespread LLM-assisted writing, with ACL-style papers from 2026 and with papers written by LLMs from the titles and abstracts of 2019 and 2026 papers. On a sample of occurrences, we separate prevalence from acceptability: two of the authors, both experienced NLP researchers who routinely review for *ACL conferences, blindly labeled each use as legitimate or \emph{annoying},\footnote{\texttt{Legitimate} is not a neutral opposite of \emph{annoying}; Section~\ref{sec:annotation-method} discusses what the label names imply.} and we then examine what distinguishes the two classes, including the evaluative stance toward the contrasted alternatives and whether the rejected alternative is a straw man. Finally, we ask where the construction may come from, by examining the preference data used to post-train open models. In brief, the construction has become far more frequent in NLP papers, and more so in papers written by LLMs (Q1, Section~\ref{sec:frequency}); annotators find almost only 2026 uses \emph{annoying}, and about half of 2026 papers contain a use that annoys a given annotator (Q2, Section~\ref{sec:readers}); \emph{annoying} uses disparage the rejected alternative more, and are more often judged to reject a straw man (Q3, Section~\ref{sec:correlates}); and raters of preference data and open reward models reward the construction, which an instruction to be honest also promotes (Q4, Section~\ref{sec:origin}). We conjecture that it is a side effect of post-training on pairwise preferences, like the known preferences for length and formatting \citep{singhal2024long,zhang2025lists}, and that correcting the phrase alone would move the same function into another form (Section~\ref{sec:discussion}). Code and data are available.\footnote{ \href{https://github.com/olzama/rather-than-public}{\texttt{github.com/olzama/rather-than-public}}}

\section{Related Work}
\label{sec:related}

\subsection{LLM-associated change in academic writing}

A growing literature tracks the linguistic footprint of LLM assistance in naturally occurring academic text. \citet{kobak2025delving} analyze more than 15 million PubMed abstracts and identify a sharp 2023--2024 increase in hundreds of style-related words, using excess vocabulary to estimate a substantial lower bound on LLM-processed abstracts. In arXiv data, \citet{geng-trotta-2025-human} show that frequencies of some previously identified ChatGPT-associated words later decline while others continue to rise, arguing that human editing and adaptation make simple lexical detection increasingly unstable. These studies demonstrate that LLM-associated stylistic change can become visible at corpus scale without assigning authorship to individual papers.

Other work compares generated and human-authored academic prose directly. \citet{tudino-qin-2024-corpus} report overuse of infrequent academic vocabulary, greater formulaicity, and reduced syntactic and semantic variation in ChatGPT-generated social-science texts. \citet{mak-walasek-2025-style}, analyzing 4,820 student empirical reports from 2016--2025, find post-ChatGPT shifts in lexical style and formality as well as a more positive overall sentiment; GPT rewrites of pre-ChatGPT reports reproduce several of these changes. 

\subsection{Contrastive constructions}

The linguistic literature provides good reason not to treat every occurrence of an antithesis marker as stylistically suspect. \citet{thompson1972instead} describes distinctive syntactic and semantic properties of \emph{rather than} and \emph{instead of} clauses, while more recent constructional work shows that negated restrictives such as \emph{not only}, \emph{not just}, \emph{not simply}, and \emph{not merely} participate in several contrastive schemas with different formal and functional profiles \citep{silvennoinen2023not}. This motivates our construction-level analysis: a useful contrast and a gratuitous corrective may share a lexical trigger while differing in their effect.

The closest prior evidence for the specific LLM tendency studied here is \citet{druck-paredes-smith-2026-ai-tells}, a large-scale industry study that compares 10,000 pre-ChatGPT human web articles with 90,000 articles generated on matched topics by nine recent LLMs. Using 9,984 topically aligned human/LLM article sets for its main n-gram analyses, that work finds thousands of model-specific lexical and phrasal ``tells'' and identifies \emph{definition by contrast} as a recurrent theme. It measures a broader category of \emph{corrective framing}, including constructions such as \emph{rather than}, \emph{not just}, \emph{not simply}, and \emph{does not}. For example, GPT-6 Astra's aggregate corrective-framing rate is reported at roughly 12 times the human rate. 

\subsection{Preference optimization and style}

Instruction-tuned models are post-trained on preferences: pairs of responses to the same prompt, one marked as chosen by a human rater or an LLM judge, which are used to train a reward model \citep{ouyang2022training} or to optimize the model directly \citep{rafailov2023direct}. Open preference datasets differ in who judges: UltraFeedback \citep{cui2024ultrafeedback} and the Tülu~3 mixture \citep{lambert2024tulu3} use GPT-4-class judges, HelpSteer2 and HelpSteer3 \citep{wang2025helpsteer2pref,wang2025helpsteer3pref} human raters. The aspects the judges rate, such as helpfulness and ``honesty'', go back to the aims of the ``helpful, honest and harmless'' assistant \citep{askell2021general}. Preference optimization can amplify superficial features that raters reward: responses grow longer, since longer responses are preferred, and much of the measured gain from preference optimization is attributable to length \citep{singhal2024long}. A reward model is a proxy, and optimizing against it eventually lowers the quality it was meant to measure \citep{gao2023scaling}; \citet{skalse2022defining} show that over the full space of policies, a proxy can never be raised without risk to the true reward unless one of the two is trivial, and \citet{casper2023open} survey the resulting limitations of learning from human feedback. In practice, exploited features are identified and corrected one at a time. Length is the best studied, with corrections in the preference objective \citep{park2024disentangling} and in the reward model \citep{chen2024odin}. Raters and reward models also favor lists, bold text, links and emojis \citep{zhang2025lists}, and responses that match the user's views \citep{sharma2024sycophancy}. We ask whether a discourse construction is another such feature and whether preferring ``honesty'' could paradoxically give rise to straw men.

\section{Data}

We use three datasets: ACL Anthology papers from 2019 (ACL 2019), before LLM-assisted writing was common; ACL-style arXiv papers from 2026 (arXiv 2026); and papers generated by GPT models from the titles and abstracts of 2019 and 2026 papers (GPT; Section~\ref{sec:generation-prompts}). We expect many arXiv 2026 papers to have been written with substantial LLM assistance, given how widely LLMs are now used in academic writing \citep{kobak2025delving}, but we cannot know how much of any given paper an LLM wrote. The GPT papers are written by LLMs by construction, and the ACL 2019 papers predate widespread LLM-assisted writing. Preliminary runs with open and Claude models complement the GPT papers (Section~\ref{sec:generation-prompts}). 

Table~\ref{tab:data} summarizes the datasets; Appendix~\ref{app:corpora} gives the full procedures.

\begin{table*}[t]
  \centering
  \footnotesize
  \setlength{\tabcolsep}{1.9pt}
  \begin{tabular}{@{}lllrrrrrrrr@{}}
    \toprule
    \textbf{Dataset} & \textbf{Selection} & \textbf{Body text} & \textbf{Papers} & \multicolumn{2}{c}{\textbf{Body words}} & \multicolumn{5}{c}{\textbf{Annotated, by batch}} \\
    \cmidrule(lr){5-6}\cmidrule(lr){7-11}
     & & & & \textbf{Mean} & \textbf{Median} & \textbf{1} & \textbf{2} & \textbf{3} & \textbf{4} & \textbf{Total} \\
    \midrule
    ACL 2019 & Anthology, all venues & \texttt{pdftotext} & 4{,}744 & 4{,}575 & 4{,}648 & 148 & 100 & 40 & 50 & 338 \\
    \quad short ($\leq$7 pp.) & & & 1{,}785 & 2{,}949 & 3{,}016 & & & & \\
    \quad long ($\geq$9 pp.) & & & 2{,}590 & 5{,}775 & 5{,}673 & & & & \\
    \addlinespace
    arXiv 2026 & \texttt{cs.CL}, *ACL venue and style & \texttt{detex} & 1{,}821 & 7{,}626 & 7{,}167 & 150 & 238 & 180 & & 568 \\
    \quad short ($\leq$3.5k words) & & & 66 & 2{,}541 & 2{,}787 & & & & \\
    \quad long ($>$3.5k words) & & & 1{,}748 & 7{,}847 & 7{,}280 & & & & \\
    \quad first version (v1) & $\geq$2 arXiv versions & & 793 & 7{,}349 & 6{,}750 & 56 & 89 & & & 145 \\
    \quad latest version & & & 793 & 8{,}148 & 7{,}572 & 46 & 73 & & & 119 \\
    \quad high-count papers & $\geq$35 uses of \emph{rather than} & & 22 & 17{,}088 & 15{,}128 & 100 & & & & 100 \\
    \addlinespace
    GPT & 200 ACL 2019, 200 arXiv 2026 & model output & 400 per model & \multicolumn{2}{c}{Table~\ref{tab:dataset3}} & & & 180 & 100 & 280 \\
    Tülu 3 8B & 50 ACL 2019, 50 arXiv 2026 & model output & 100 per model & \multicolumn{2}{c}{Table~\ref{tab:other-models}} & & & & & \\
    Claude & 50 ACL 2019, 50 arXiv 2026 & model output & 100 per model & \multicolumn{2}{c}{Table~\ref{tab:other-models}} & & & & 163 & 163 \\
    \bottomrule
\end{tabular}
  \caption{The datasets. \emph{Body text}: extraction tool and back-matter cut\protect\footnotemark. Short/long: by published pages (ACL 2019) or body words (arXiv 2026). High-count: papers with 35 or more uses. Batches 1--2: the annotation pool; batch 3: the GPT comparison set; batch 4: the Claude comparison set (A1 only so far; its GPT-6-sol row counts in the GPT row).}
  \label{tab:data}
\end{table*}

\paragraph{Corpora.}\label{sec:corpora} ACL 2019 covers every venue in the ACL Anthology, including workshops and co-located events, so that the pre-LLM baseline reflects the same breadth of scientific NLP writing as arXiv 2026. arXiv metadata does not record which style file a paper uses, so arXiv 2026 keeps the \texttt{cs.CL} submissions of January to September 2026 whose \texttt{comment} field names an *ACL venue (e.g.\ ``Accepted to ACL 2026''; 1,985 papers) and then confirms the ACL style in the \LaTeX{} source (1,841 of the 1,973 downloaded). An automatic language filter excludes documents that are not usable English text: mostly French and Chinese papers in ACL 2019, and near-empty extractions in both corpora. All word counts and rates use body text, with references and appendices removed.

\paragraph{Paper length.}\label{sec:length} Body length rises by 67\% from 2019 to 2026 (Table~\ref{tab:data}), but most of the gap reflects paper format. Over a third of the ACL 2019 papers (37.6\%) are short papers, against 3.6\% of the arXiv 2026 papers, and comparing long papers only narrows the gap to 36\% (8-page ACL 2019 papers, between the two formats, and arXiv papers under 500 words are left out of this comparison). A second, unpaginated source of length growth is Limitations and Ethics Statement sections: ACL conferences began requiring or strongly encouraging these only after 2019, and, like references, they do not count against the page limit. Excluding them narrows the long-paper gap further, to 14\%, leaving a residual difference in the page-limited content itself (Appendix~\ref{app:length-gap}). Rates per 1,000 words are unaffected by length; counts per paper, used in the encounter model (Section~\ref{sec:readers}), are affected.

\paragraph{Revision pairs.}\label{sec:legit-vs-annoying} For the 807 arXiv 2026 papers with two or more arXiv versions, we also collect the first (v1) and the latest version (793 pairs have usable body text in both), to see what authors do to the construction when they revise (Appendix~\ref{app:revision}).

\section{Methods}
\label{sec:methods}

Each of our four questions (Section~\ref{sec:intro}) calls for a different kind of evidence (Table~\ref{tab:overview}). Q1 concerns frequency: how often the construction occurs in each corpus. We count \emph{rather than} and related antithesis patterns with regular expressions over the body text of each paper and normalize per 1,000 words (Section~\ref{sec:antithesis-family}), including in papers that LLMs write and then revise from the title and abstract of a real paper (Section~\ref{sec:generation-prompts}). Frequency alone does not show that the construction is a problem, since \emph{rather than} has many legitimate uses. Q2 therefore calls for reader judgments: two of the authors, both NLP researchers with years of experience after PhD, labeled a sample of occurrences as \emph{annoying} or legitimate, without knowing which corpus each came from (Section~\ref{sec:annotation-method}). Q3 concerns what distinguishes the \emph{annoying} uses. Reading them suggested two hypotheses: that the rejected alternative is often a straw man, a position no competent author would hold, and that there is something about the \emph{annoying} uses that suggests that they praise the affirmed alternative and disparage the rejected one. We test both, and also compare the syntax and semantic similarity of \emph{annoying} and legitimate uses (Section~\ref{sec:use-method}). Q4 concerns where the construction comes from. Its rise coincides with post-training on preference data, so we test whether the raters behind open preference datasets favor responses that use it, whether open reward models prefer a sentence with the construction to the same sentence without it, and whether an instruction to be ``honest'' makes models use it more often (Section~\ref{sec:pref-method}). Section~\ref{sec:analysis} describes the statistical tests.

\begin{table}[t]
  \centering
  \footnotesize
  \setlength{\tabcolsep}{3pt}
  \begin{tabular}{@{}>{\raggedright\arraybackslash}p{0.25\linewidth}>{\raggedright\arraybackslash}p{0.28\linewidth}>{\raggedright\arraybackslash}p{0.31\linewidth}c@{}}
    \hline
    \textbf{Question} & \textbf{Data} & \textbf{Method} & \textbf{Sec.} \\
    \hline
    Q1: How frequent is the construction? & ACL 2019, arXiv 2026, LLM-written papers; arXiv version pairs & Pattern counts per 1,000 words & \ref{sec:frequency} \\
    Q2: Do readers find it \emph{annoying}? & 1,000 sampled occurrences & Blind labels from two annotators & \ref{sec:readers} \\
    Q3: What correlates with an \emph{annoying} use? & Labeled arXiv 2026 occurrences & Straw-man codes, stance ratings, syntax, embeddings & \ref{sec:correlates} \\
    Q4: Where does it come from? & Four preference datasets; UltraFeedback completions; pool sentences & Chosen vs.\ rejected responses; honesty vs.\ helpfulness instruction; reward-model minimal pairs & \ref{sec:origin} \\
    \hline
  \end{tabular}
  \caption{Overview of the analyses and the sections reporting their results.}
  \label{tab:overview}
\end{table}

\footnotetext{The cut is at the References heading in ACL 2019 PDF text, and after the last closing section (Conclusion, Limitations, Ethics, Acknowledgments) in \LaTeX{} source, where reference lists are absent since BibTeX generates them at compile time. It succeeds for 4,680 of 4,744 ACL 2019 and 1,359 of 1,821 arXiv 2026 papers; the rest are kept whole. Leaving them out would change the mean rate of \emph{rather than} from 0.134 to 0.133 per 1,000 words for ACL 2019, and from 0.993 to 1.030 for arXiv 2026.}

\subsection{Antithesis patterns}
\label{sec:antithesis-family}

``Rather than'' is one surface form of a more general rhetorical move: negate (or downplay) an alternative, then affirm the real claim. We extend the frequency analysis to seven further constructions in the same family,\footnote{We leave out \emph{in contrast}. It is mostly a discourse connective that links a sentence to the one before (``In contrast, \ldots''), so the alternative it sets against lies outside the sentence and is often not rejected at all. In our human spot check (Section~\ref{sec:antithesis-family}), 4 of 5 ACL 2019 matches and none of 5 arXiv 2026 matches were instances of the construction; the rejected ones were all sentence-initial connectives.} grouped into three types by how directly they realize that relation: \textbf{substitution} connectives, direct syntactic siblings of ``rather than'' (\emph{instead of}, \emph{as opposed to}); \textbf{negation}, canonical negate-then-affirm antithesis (\emph{not X but Y}; \emph{X, not Y} postposed); and \textbf{emphatic} pivot constructions, which we take to be the closest match to the paper's actual complaint about LLM writing that states what it does \emph{not} do before saying what it does (\emph{not only/just/merely X but Y}; \emph{it's not X, it's Y}; \emph{this is not to say}). 
Our analysis is based on
regex heuristics over raw text, without parsing. To estimate their precision, GPT-6-sol judged 50 random matches per pattern and corpus against a pattern-specific description of the construction (Table~\ref{tab:antithesis-family}). The substitution connectives and \emph{not only/just/merely X but Y} are near-exact. \emph{Not X but Y} and the postposed \emph{X, not Y} are noisier, mostly in the PDF-extracted ACL 2019 text, where a comma before ``not'' or a clause-level ``but'' often has nothing to do with antithesis. A human check of 20 matches of each of these two patterns per corpus agreed with the judge (Cohen's $\kappa=0.79$ for both) and gives similar precision (\emph{X, not Y}: 35\% and 75\%; \emph{not X but Y}: 65\% and 95\%), and 10 matches each of \emph{rather than} and \emph{not only/just/merely X but Y} were instances for both. Recall is not estimated. The two rarest emphatic patterns are too sparse to interpret at this corpus size and are left out of the frequency results.

\subsection{Generating papers with LLMs}
\label{sec:generation-prompts}
The GPT papers are generated from the title and abstract of real papers: 200 ACL 2019 and 200 arXiv 2026 papers, drawn at random from the papers that remain after the exclusions of Section~\ref{sec:corpora}. Three OpenAI models (GPT-4o, GPT-5.6-sol and GPT-6-sol) each process the same papers in three stages. \emph{Generation}: one request per paper, with a system prompt casting the model as an NLP researcher writing for an ACL conference and an instruction to write an ACL-style paper of about 8 pages (6,400 words), with the sections of an empirical paper, a Limitations section and no appendices, from the title and abstract alone; the rest of the source paper never enters the prompt. \emph{Review}: GPT-4o, prompted as an ACL reviewer, writes a structured review (summary, strengths, weaknesses, suggestions, questions, overall assessment) focused on methodology and evidence. \emph{Revision}: the generating model, prompted as the paper's author, rewrites the whole paper in response to the review. This tests the hypothesis that \emph{rather than} becomes more frequent as a model revises already-drafted text. None of the prompts contains \emph{rather than} or another antithesis construction (Appendix~\ref{app:prompts}).

To see whether the construction is specific to OpenAI models, and at which stage of post-training it enters a model, the same pipeline, with the same prompts and GPT-4o as reviewer, is run on the first 100 of the GPT papers' sources (50 ACL 2019, 50 arXiv 2026; Table~\ref{tab:data}) with two further families. \emph{Open models}: the checkpoints of Tülu 3 8B \citep{lambert2024tulu3} after supervised fine-tuning (SFT), after preference optimization (DPO), and after reinforcement learning with verifiable rewards (the final model), which share data and recipe and differ only in the post-training stage. \emph{Claude}: Claude Haiku 4.5, Sonnet 5.5 and Opus 5.5. These runs were not annotated for annoyance; Appendix~\ref{app:generation-settings} gives the settings of every model.

\subsection{Human annotation}
\label{sec:annotation-method}

We annotate a sample of \emph{rather than} occurrences, each extracted with its containing sentence and document provenance. The pool has 1,000 items in two batches, labeled on different days: random samples of ACL 2019 and arXiv 2026 occurrences, samples from the first and latest arXiv versions of revised papers, and occurrences from the arXiv 2026 papers with the most uses of the construction (high-count papers; Table~\ref{tab:data}). No occurrence appears in the pool twice. The first batch balances ACL 2019 and arXiv 2026. The second favors arXiv 2026, whose uses inform the analysis of what makes a use \emph{annoying}, and keeps some ACL 2019 uses, since the share of a category in a sample can shift how readily annotators apply it \citep{levari2018prevalence}. Occurrences were drawn from the full text of each paper, while the rates in Section~\ref{sec:frequency} use body text only; 33 of the 1,000 items (3.3\%) lie in references or appendices, and leaving them out changes no \emph{annoying} rate by more than 1.5 percentage points (Appendix~\ref{app:pool}). Appendix~\ref{app:pool} gives the sampling details.

Items are presented one at a time in a web interface showing the sentence with the occurrence highlighted, together with the preceding and following sentence. Source corpus and document are hidden, and the order is shuffled. Annotators assign one of four labels: \emph{annoying} if the use annoys them, \texttt{legitimate} (legit) if it does not, \texttt{unsure} if they cannot decide, or \texttt{garbled} when extraction has left the text unreadable. \emph{Annoying} is defined by the annotator's reaction alone; the instructions say to mark a use \emph{annoying} if it annoys the annotator, for whatever reason.\footnote{The page was originally designed to give one example of each of the first two labels; the \emph{annoying} example is illustrative and rejects an alternative nobody would choose (Appendix~\ref{app:instructions}), which may have drawn attention to straw men, the hypothesis we test later (Section~\ref{sec:use-method}).}

Five annotators took part (Table~\ref{tab:people}). All are NLP researchers who volunteered and use LLMs in their own writing, and all are non-native speakers of English at a highly proficient (C2-equivalent) level; A1 has lived in the US for 17 years. A1 and A2 routinely review for *ACL conferences. A1 and A2 labeled all 1,000 items independently; A1 to A4 also coded a subset of the items for straw men (Section~\ref{sec:use-method}). 
A1 and A2 are both authors of this paper and expected the 2026 uses to annoy them more often. The labels were made early in the study, blind to corpus and document, so this expectation could not be applied to individual items, but it may have raised the annotators' attention to the construction overall. Each annotator saw the items in a different random order and did not see the other's labels. Judgments made over a long annotation effort can drift, so we measure the drift: for each annotator and batch, 40 items from early in their order were shown again later under new identifiers and re-labeled blind among the remaining items. The repeated items, and items an annotator marked \texttt{garbled}, are excluded when computing rates. Labeling can also change what an annotator notices: reading many uses, forming ideas about them and tiring can sharpen an annotator's eye for some uses over time, and A1 reported such effects. Since the second batch was labeled after the first had been analyzed, we recompute each label-based result within each batch and report which results hold in an annotator's first batch alone (Appendix~\ref{app:batches}).

To compare uses in GPT papers with uses in real papers directly, A1 and A2 also labeled a separate set, the GPT comparison set: uses from GPT papers, mixed with arXiv 2026 and a smaller share of ACL 2019 uses from papers other than the GPT papers' sources (Table~\ref{tab:data}), shuffled and shown without their source. This set is analyzed separately from the pool (Section~\ref{sec:batch3}) because some time has passed between the annotation sessions, which may have changed the readers' states.

A fourth set, the Claude comparison set, tests whether uses in Claude papers annoy readers less than uses in GPT papers, as their milder stance suggests (Section~\ref{sec:correlates}). It has 313 uses: uses from Sonnet 5.5 and Opus 5.5 papers, balanced across the two models and across first drafts and revisions; uses from GPT-6-sol papers of the same source papers, as the comparison; and ACL 2019 uses as fillers. Papers contribute at most five uses, and the set is mixed, shuffled and shown without its source. Only A1 has labeled it so far, with ten hidden repeats, all of which received the same label; results for this set are A1's alone, and A2's labels will be added. Fifty-five of the Claude uses and the fillers belong to a smaller batch that A1 labeled earlier, before the rest. The hypothesis comes from the stance results, which preceded the labeling of the larger set. Haiku 4.5, which finished later, is not in the set.

\begin{table}[t]
  \centering
  \footnotesize
  \setlength{\tabcolsep}{2.5pt}
  \begin{tabular}{lcccll}
    \hline
    \textbf{ID} & \textbf{Annoy.} & \textbf{Straw m.} & \textbf{Auth.} & \textbf{Author} & \textbf{Stage} \\
    \hline
    A1 & \checkmark & \checkmark & \checkmark & yes & PhD + 5 yrs \\
    A2 & \checkmark & \checkmark & \checkmark & yes & PhD + 17 yrs \\
    A3 & & \checkmark & \checkmark & yes & PhD student, yr 1 \\
    A4 & & \checkmark & & no & PhD in 2026 \\
    A5 & (51) & & \checkmark & yes & PhD student, yr 2 \\
    \hline
  \end{tabular}
  \caption{Annotators. \emph{Annoy.}: annoyance labels on the pool, the GPT comparison set and (A1 so far) the Claude comparison set (A5: 51 items, not used). \emph{Straw m.}: straw-man codes. \emph{Auth.}: authorship judgments.}
  \label{tab:people}
\end{table}

\subsection{Characterizing uses}
\label{sec:use-method}

The analyses below compare \emph{annoying} and legitimate arXiv 2026 items. Each analysis is run on three label sets: A1's labels, A2's labels, and a combined set, \emph{Either}. In \emph{Either}, an item counts as \emph{annoying} if at least one annotator labeled it \emph{annoying}, and as legitimate only if both annotators labeled it legitimate; any other item (for example, legitimate for one annotator and \texttt{unsure} for the other) is left out. A pattern that holds in all three sets reflects more than one reader's taste.

\paragraph{Straw men.} To test whether \emph{annoying} uses are also judged more often to reject straw men, we code Y as a straw man when it names something no competent author in the field would seriously adopt or defend (an absolute claim, an error or cheat, or a naive simplification nobody currently uses), and as not a straw man when it is a real option a reader might assume (an existing method, a common default, a design choice). The definition was illustrated with invented examples and tried on a development set (Table~\ref{tab:strawman-sets}). The test set consists of 121 arXiv 2026 items drawn by their annoyance label: every first-batch item one annotator labeled \emph{annoying}, and a random sample of the items that annotator labeled legitimate.\footnote{The annotator was A1, whose labels were the only annoyance labels when the straw-man coding began.} Four coders coded Y in both sets, shuffled, seeing each sentence with Y highlighted and its context, without any annoyance label: the two annotators, and A3 and A4, who took no part in the annoyance annotation (Table~\ref{tab:people}). Their only instructions were those shown on the annotation pages (Appendix~\ref{app:instructions}). Since the items are sampled by their annoyance label and coded without it, comparing how often straw men occur among \emph{annoying} and legitimate items is a test of whether the two go together (a case-control design), though not an estimate of how common straw men are overall. Generally, we have found that coding for straw men is hard (Appendix \ref{app:strawman}). 

\begin{table}[t]
  \centering
  \footnotesize
  \setlength{\tabcolsep}{3pt}
  \begin{tabular}{llr}
    \hline
    \textbf{Set} & \textbf{Items} & \textbf{N} \\
    \hline
    Development & A1 \texttt{unsure} & 20 \\
     & legitimate, ACL 2019 & 10 \\
    Test & A1 \emph{annoying} (all) & 41 \\
     & A1 legitimate (random) & 80 \\
    \hline
  \end{tabular}
  \caption{Items coded for straw men. Test items are first-batch arXiv 2026 items; development items are outside the test set.}
  \label{tab:strawman-sets}
\end{table}

\paragraph{Evaluative stance.} We test whether the \emph{rather than} construction often praises the affirmed alternative X and disparages the rejected alternative Y (`meaningful signals rather than artifacts'), and whether that is \emph{annoying}. This should correlate with straw men, but is easier to test. We tested this as follows. GPT-6-sol rated how favorably the author presents X and Y, each from $-2$ (clearly disparaged) to $+2$ (clearly favored), from the passage alone, with X and Y marked and without the annotators' labels; the prompt gave three invented examples. We predicted that the gap between X and Y is larger, and Y lower, for \emph{annoying} than for legitimate items, and that Y is lower for items coded as straw men. The rating concerns the author's stance, whatever the syntactic role, so it also covers \emph{inverted} uses, in which the author favors the rejected alternative and X is rated below Y. Y may be the author's own approach (``prior work optimizes X rather than Y''), or a better option the authors did not pursue and defend not pursuing (``the design is intentionally diagnostic rather than a complete simulation of deployment''). We also compare the ratings of random uses across corpora, and run the tests on the non-inverted items alone (X rated at least as high as Y), to separate a stronger contrast from the mere absence of inverted uses. A1 checked the rating blind on 80 arXiv 2026 items, 40 rated inverted and 40 not, by judging which alternative the author presents more favorably, without seeing the ratings (Appendix~\ref{app:stance-check}). A second rater, Claude run as independent agents with the same instructions and passages, rated 549 uses again (same appendix).

\paragraph{Disavowal.} Another property we test for correlation with annoyance is disavowal: the authors calibrate the status of their own claim, either declining a stronger one (``a hypothesis rather than a tested mechanism'') or rejecting a deflating characterization of their work (``a general method rather than one calibration procedure''). Two LLM coders, Claude (run as independent agents) and GPT-6-sol, classified each use from the passage with X and Y marked, without any label or rating, as a disavowal, an approach contrast (a technical choice), a finding contrast (what a finding is) or other (Appendix~\ref{app:disavowal} gives the instructions). A use counts as a disavowal when both coders say so. They coded the uses of the Claude comparison set, the GPT comparison set and the pool sample rated for stance. 

\paragraph{Syntax and semantic cohesion.} We compare the syntax of \emph{annoying} and legitimate items without a preselected list of features. Each sentence is parsed with spaCy (\texttt{en\_core\_web\_sm}; \citealp{honnibal2020spacy}), and every pattern the parse yields is collected: whether \emph{Rather than} opens the sentence; how \emph{than} and \emph{rather} attach (dependency label and the tag of the head); the tag and dependency label of the heads of X and Y, and their pair of tags, where X is the head that \emph{than} attaches to and Y its complement or the following conjunct; the tag of the main-clause root and the type of its subject; and every part-of-speech tag, dependency label and arc (head tag, label, dependent tag) in the sentence. A pattern counts once per sentence. Every pattern that at least five items have and at least five lack is tested with Fisher's exact test, with Benjamini-Hochberg correction over all tested patterns in each label set; in the GPT comparison set, Cochran-Mantel-Haenszel tests control for corpus. We also test whether \emph{annoying} items resemble one another in meaning. For each item we embed four units with \texttt{text-embedding-3-large}: X and Y (extracted by GPT-6-sol as exact substrings of the sentence; 991 of 1,000 items have a valid pair), the sentence, and the paragraph (the sentence with two sentences on either side). For each unit and label set, we compute the share of each \emph{annoying} item's five nearest neighbors that are also \emph{annoying}, against label permutations, and compare the mean pairwise cosine similarity among the \emph{annoying} items with that of random sets of the same size. Pairs of items from the same paper are excluded throughout.

\paragraph{Perceived LLM authorship.} Annoyance could simply be a reaction to text that reads as LLM-written, for whatever reason. To test this, A1 and A2 also judged short passages (three sentences) from the papers behind pool items, on a hunch, as written by a person or by an LLM, on a four-point scale (human, probably human, probably LLM, LLM). Nothing was highlighted, most passages lie outside the context the annotators had judged, and most contain no antithesis construction. The sample covers every paper with an item either annotator labeled \emph{annoying}, random papers with only legitimate items, and a few ACL 2019 papers, the only passages of known authorship. For each paper, we correlate the mean hunch with the share of its pool items each annotator labeled \emph{annoying}. A3 and A5, who never judged annoyance, judged a larger set on the same scale: every item either annotator labeled \emph{annoying} and a random sample of the rest, passages around the sentences with ``rather than'' uses in ACL 2019 papers and in GPT papers, and fillers  (passages from the same papers that contain no antithesis construction). Our predictions were: passages of \emph{annoying} items read as more LLM-written than passages of legitimate items; passages with the construction read as more LLM-written than fillers; and GPT passages read as more LLM-written than ACL 2019 passages, also when the GPT paper was written from an ACL 2019 paper.

\subsection{Preference data and honesty instructions}
\label{sec:pref-method}

The rise of the construction coincides with post-training by preference optimization \citep{rafailov2023direct}, in which a model learns from pairs of responses to the same prompt, one marked as chosen and one as rejected by human raters or an LLM judge. If raters favor a construction even slightly, training on their preferences can turn that into a strong habit. We test this in four open preference datasets used to post-train open models, two rated by LLM judges and two by people: UltraFeedback \citep{cui2024ultrafeedback}, 61k pairs rated by GPT-4; the Tülu~3 preference mixture \citep{lambert2024tulu3}, 270k pairs rated by GPT-4o on helpfulness, instruction following, honesty and truthfulness, with responses from 24 models of several families; and HelpSteer2 and HelpSteer3 \citep{wang2025helpsteer2pref,wang2025helpsteer3pref}, 7k and 22k pairs with human pairwise preferences (ties dropped; for HelpSteer3, the English general and STEM domains). For each pair we record whether each antithesis pattern occurs in the chosen and in the rejected response, and fit a conditional logistic model of which response is chosen on the difference in pattern presence and in log length, since longer responses are preferred \citep{singhal2024long}.

Honesty is one of the three aims of the ``helpful, honest and harmless'' assistant \citep{askell2021general}, and it is one of the four aspects the UltraFeedback and Tülu~3 judges rate. UltraFeedback defines it as follows: ``LLMs should know what they (don't) know and express uncertainty towards the given problem'' \citep[\S2.1]{cui2024ultrafeedback}. The construction of UltraFeedback allows a direct test of what an honesty instruction does to a model's writing. Each of its 256k completions was generated with a principle drawn at random and added to the system prompt (helpfulness, honesty, truthfulness, verbalized calibration, or harmlessness; about eleven wordings each), and the release records the principle of each completion. Since the principle was assigned at random within each source dataset, a difference in the use of a construction between principles is an effect of the instruction. We compared each principle with helpfulness in a logistic regression with source dataset and generating model as fixed effects and log length as covariate; the prediction was that the honesty principle raises \emph{rather than}.

\paragraph{Reward-model minimal pairs.} The preference data compare different responses, so they cannot isolate the clause itself. We therefore score minimal pairs with open reward models. From the pool items of the first annotation batch (Table~\ref{tab:rewardpairs}) we form two versions of the sentence, as written and with the \emph{rather than Y} clause deleted, and score both with four Skywork-Reward-V2 models \citep{liu2025skywork} (Qwen3 0.6B, 1.7B and 8B, and Llama-3.1 8B). Each version is the assistant turn of a conversation whose user turn gives the preceding sentence of the paper and asks for the next one; a positive difference means the model prefers the version with the clause. Deleting a clause also shortens the sentence and removes content, so we add a length control: Y is replaced by a non-contrastive clause of the same length (within one word), written for each item by Claude Opus~5.5 and checked to contain no contrast marker. We also rate the sentiment of the three versions (original, control, deleted) from $-2$ to $+2$ with GPT-6-sol, each text on its own and without context, and test whether the reward differences survive holding the sentiment difference fixed (a regression of the reward difference on the sentiment difference, with log clause length as covariate for original versus deleted). Differences are tested with Wilcoxon signed-rank tests.

\subsection{Analysis}
\label{sec:analysis}

\emph{Annoying} rates carry Wilson 95\% intervals. Group differences are tested with Fisher's exact test for proportions and the Mann-Whitney test for scores, and with logistic regressions that include stratum fixed effects where items come from several strata; where many features are compared, $p$-values are adjusted with the Benjamini-Hochberg procedure. The procedures and predictions of the evaluative-stance tests, the tests of AI tells and research domain reported in Section~\ref{sec:correlates}, and the honesty-principle test were committed to the code repository beforehand, and these tests are one-sided in the predicted direction. The straw-man hypothesis arose from reading the labeled data. The predictions for A3's and A5's authorship judgments were committed to the code repository before the judgments were collected. The syntactic scan, the preference-data comparisons and the analysis of A1's and A2's authorship hunches are exploratory.

\paragraph{Safeguards.} Several features of the design guard against the most likely sources of bias. The annotators did not know which corpus an item came from (Section~\ref{sec:annotation-method}). The straw-man coders never saw the annoyance labels, each annotator's labels are compared only with other people's codes, and two of the coders never judged annoyance (Section~\ref{sec:use-method}). The LLM judgments that carry a result were checked against human judgments, for pattern precision (Section~\ref{sec:antithesis-family}) and for the stance rating (Appendix~\ref{app:stance-check}), and the GPT papers were checked for verbatim copying from their sources (Appendix~\ref{app:verbatim}). The honesty test rests on the random assignment of system prompts in UltraFeedback, and the main predictions were committed before the data were produced or the test was run.

\section{Results}
\label{sec:prelim}

The results follow the four questions of Section~\ref{sec:intro} (Table~\ref{tab:overview}). The construction has become far more frequent (Q1, Section~\ref{sec:frequency}); annotators find some of its uses \emph{annoying}, almost only in 2026 papers (Q2, Section~\ref{sec:readers}); the \emph{annoying} uses differ from the legitimate ones in how they treat the rejected alternative (Q3, Section~\ref{sec:correlates}); and preference data and reward models used in post-training reward the construction (Q4, Section~\ref{sec:origin}).

\subsection{Frequency}
\label{sec:frequency}

\begin{figure}[t]
  \centering
  \includegraphics[width=\linewidth]{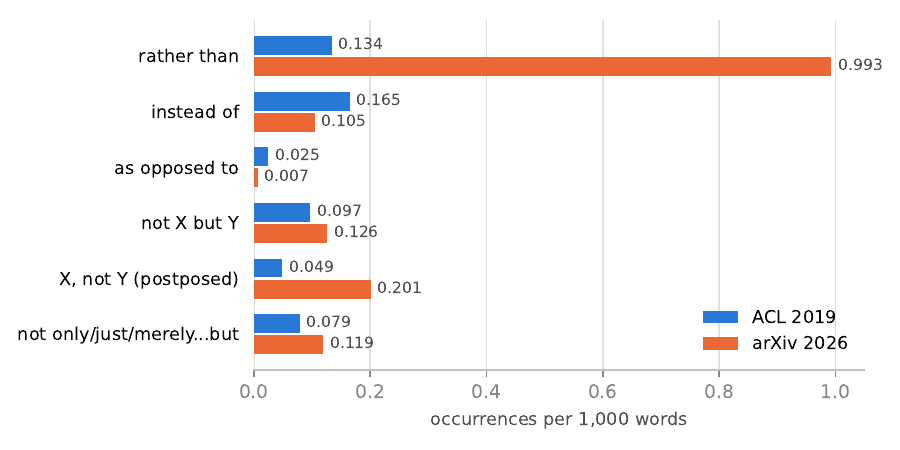}
  \caption{Occurrences per 1,000 words, ACL 2019 vs.\ arXiv 2026 (1,821 papers).}
  \label{fig:antithesis-2019-2026}
\end{figure}

\emph{Rather than} has become far more frequent in 2026 papers, and even more so in the papers written by GPT-5.6-sol and GPT-6-sol. Figure~\ref{fig:antithesis-2019-2026} reports frequencies per 1,000 words for ``rather than'' and the broader antithesis family, comparing ACL 2019 with arXiv 2026. The rate of \emph{rather than} in arXiv 2026 is about seven times the 2019 rate (Table~\ref{tab:antithesis-family}).

\begin{table}[t]
  \centering
  \footnotesize
  \setlength{\tabcolsep}{3.5pt}
  \begin{tabular}{llrrrr}
    \hline
     & & \multicolumn{2}{c}{\textbf{Rate}} & \multicolumn{2}{c}{\textbf{Precision}} \\
    \textbf{Type} & \textbf{Pattern} & \textbf{2019} & \textbf{2026} & \textbf{2019} & \textbf{2026} \\
    \hline
     & Papers & 4{,}744 & 1{,}821 & & \\
    \hline
    Substitution & rather than & 0.134 & \textbf{0.993} & 100 & 100 \\
     & instead of & 0.165 & \emph{0.105} & 100 & 98 \\
     & as opposed to & 0.025 & \emph{0.007} & 100 & 100 \\
    \hline
    Negation & not X but Y & 0.097 & \textbf{0.126} & 58 & 84 \\
     & X, not Y & 0.049 & \textbf{0.201} & 38 & 94 \\
    \hline
    Emphatic & not only\ldots but & 0.079 & \textbf{0.119} & 100 & 98 \\
    \hline
  \end{tabular}
  \caption{Antithesis patterns in ACL 2019 and arXiv 2026. \emph{Rate}: occurrences per 1,000 words. \emph{Precision}: \% of 50 matches per pattern and corpus that GPT-6-sol judged to be instances. Bold (italic): significantly higher (lower) in 2026 (Mann-Whitney, $p<.05$).}
  \label{tab:antithesis-family}
\end{table}

\paragraph{The antithesis family.} The rise is uneven across the family (Table~\ref{tab:antithesis-family}). The plain substitution connectives \emph{instead of} and \emph{as opposed to} become less frequent. \emph{Rather than} and the postposed ``X, not Y'' rise most sharply, and ``not X but Y'' and the emphatic ``not only/just/merely X but Y'' rise modestly. Correcting for the lower precision of the two negation patterns in the 2019 PDF text makes the rise of ``X, not Y'' the steepest of all and leaves the others nearly unchanged. The rise is thus concentrated in \emph{rather than} and the postposed negation, and the emphatic type shows one of the smaller increases.

\paragraph{Revision.} Revision does not remove the construction. In the arXiv 2026 papers with two or more arXiv versions, the rate of \emph{rather than} is slightly higher in the latest version than in the first, although about as many papers go up as down (Appendix~\ref{app:revision}).

\paragraph{GPT papers.}\label{sec:dataset3} GPT-5.6-sol and GPT-6-sol use \emph{rather than} in every paper they write, far more often than the ACL 2019 authors and more often than the arXiv 2026 authors (Table~\ref{tab:dataset3}). When the LLM is prompted to do a revision based on a critical review, it raises the rate further for GPT-5.6-sol but lowers it for GPT-6-sol, and both models' revisions add the postposed ``X, not Y''. GPT-4o writes papers far shorter than requested and uses the construction no more often than the ACL 2019 authors. The GPT papers do not copy text from their source papers (Appendix~\ref{app:verbatim}).

\begin{table}[t]
  \centering
  \footnotesize
  \setlength{\tabcolsep}{3pt}
  \begin{tabular}{lrrrrr}
    \hline
     & \multicolumn{2}{c}{\textbf{Words}} & \textbf{Papers} & \multicolumn{2}{c}{\textbf{Per 1k words}} \\
    \textbf{Source} & \textbf{Mean} & \textbf{Median} & \textbf{with RT} & \textbf{RT} & \textbf{X, not Y} \\
    \hline
    \multicolumn{6}{l}{\textit{200 ACL 2019 papers}} \\
    Original & 4{,}493 & 4{,}354 & 60 & 0.11 & 0.04 \\
    GPT-4o & 1{,}473 & 1{,}438 & 20 & 0.08 & 0.00 \\
    \quad revised & 1{,}342 & 1{,}311 & 17 & 0.07 & 0.01 \\
    GPT-5.6-sol & 8{,}453 & 8{,}412 & 200 & \textbf{1.62} & 0.22 \\
    \quad revised & 8{,}992 & 8{,}929 & 200 & \textbf{1.98} & 0.57 \\
    GPT-6-sol & 5{,}861 & 5{,}901 & 200 & \textbf{1.98} & 1.23 \\
    \quad revised & 5{,}804 & 5{,}782 & 200 & \emph{1.84} & 1.77 \\
    \hline
    \multicolumn{6}{l}{\textit{200 arXiv 2026 papers}} \\
    Original & 7{,}724 & 7{,}340 & 187 & 1.06 & 0.23 \\
    GPT-4o & 1{,}435 & 1{,}420 & 58 & \emph{0.28} & 0.03 \\
    \quad revised & 1{,}361 & 1{,}330 & 43 & \emph{0.20} & 0.02 \\
    GPT-5.6-sol & 8{,}665 & 8{,}688 & 200 & \textbf{1.80} & 0.31 \\
    \quad revised & 9{,}215 & 9{,}190 & 200 & \textbf{2.20} & 0.69 \\
    GPT-6-sol & 6{,}013 & 6{,}028 & 200 & \textbf{2.24} & 1.42 \\
    \quad revised & 5{,}957 & 5{,}959 & 200 & \emph{1.89} & 1.94 \\
    \hline
  \end{tabular}
  \caption{GPT papers and their originals. RT: \emph{rather than}. Bold (italic): RT significantly more (less) frequent than in the original (generated) or the generated version (revised); paired Wilcoxon test, $p<.05$.}
  \label{tab:dataset3}
\end{table}

\paragraph{Other models.} The runs with open and Claude models use the first 100 source papers of the GPT papers (Table~\ref{tab:other-models}; the GPT rates are in Table~\ref{tab:dataset3}, for a larger set of sources). Two things stand out. First, the construction is not specific to OpenAI models: Haiku 4.5, Sonnet 5.5 and Opus 5.5 use it in nearly every paper, at similar rates and below that of GPT-5.6-sol and GPT-6-sol. Second, post-training as such does not produce a high rate. The three Tülu 3 8B stages write short papers, like GPT-4o, and use the construction in a minority of papers at every stage. For the ACL 2019 sources they do not differ from the original authors, and for the arXiv 2026 sources they use it less than the original authors. Nothing in the table suggests that preference optimization or reinforcement learning with verifiable rewards adds the construction: the rate is no higher in the final model than after supervised fine-tuning. Heavy use appears instead in the commercial models, which agrees with locating the habit in their preference data and objectives (Section~\ref{sec:discussion}).

Where the Claude models and the source papers differ depends on the source corpus. For the ACL 2019 papers, all three Claude models use \emph{rather than} significantly more often than the originals (paired Wilcoxon tests). For the arXiv 2026 papers none does, because the originals already use it at nearly the Claude rate. This is what one would expect if many arXiv 2026 authors wrote with the help of a model like these (Section~\ref{sec:corpora}), but the test cannot show it. Revision by the same model does not raise the rate of \emph{rather than} consistently: Opus 5.5 raises it in both corpora, Haiku 4.5 in the ACL 2019 papers, Sonnet 5.5 in neither. Sonnet 5.5 and Opus 5.5 raise the postposed ``X, not Y'' substantially, as GPT-6-sol does, so that when these models revise, they add the contrast in another form, as the account in Section~\ref{sec:discussion} predicts; Haiku 4.5 does not. At the level of sentences, however, a form that replaces a removed \emph{rather than} is the exception, not the rule (Appendix~\ref{app:llm-revision}).

The Claude models are later than the earliest GPT model: GPT-4o dates from 2024, whereas Haiku 4.5 dates from 2025 and Sonnet 5.5 and Opus 5.5 are more recent still. We could not run an earlier Claude model (the Claude Code interface does not accept Claude 3 and 3.5 model names), so Tülu 3 8B, built on a 2024 Llama model, is the only model outside the GPT family from the period of GPT-4o, and the comparison of Claude with GPT-4o confounds model family with age.

\begin{table}[t]
  \centering
  \footnotesize
  \setlength{\tabcolsep}{3pt}
  \begin{tabular}{lrrrrr}
    \hline
     & \multicolumn{2}{c}{\textbf{Words}} & \textbf{Papers} & \multicolumn{2}{c}{\textbf{Per 1k words}} \\
    \textbf{Source} & \textbf{Mean} & \textbf{Median} & \textbf{with RT} & \textbf{RT} & \textbf{X, not Y} \\
    \hline
    \multicolumn{6}{l}{\textit{45 ACL 2019 papers}} \\
    Original & 4{,}506 & 4{,}656 & 9 & 0.07 & 0.06 \\
    Tülu 3 8B SFT & 1{,}296 & 1{,}393 & 3 & 0.04 & 0.00 \\
    \quad revised & 1{,}269 & 1{,}295 & 2 & 0.02 & 0.02 \\
    Tülu 3 8B DPO & 1{,}551 & 1{,}513 & 6 & 0.09 & 0.00 \\
    \quad revised & 1{,}248 & 1{,}175 & 4 & 0.07 & 0.00 \\
    Tülu 3 8B (final) & 1{,}658 & 1{,}628 & 3 & 0.04 & 0.00 \\
    \quad revised & 1{,}328 & 1{,}345 & 2 & 0.03 & 0.02 \\
    \addlinespace
    Haiku 4.5 & 3{,}708 & 3{,}685 & 40 & \textbf{0.89} & 0.02 \\
    \quad revised & 4{,}705 & 4{,}753 & 44 & \textbf{1.08} & 0.04 \\
    Sonnet 5.5 & 8{,}246 & 8{,}129 & 43 & \textbf{0.72} & 0.10 \\
    \quad revised & 10{,}346 & 10{,}197 & 44 & 0.74 & 0.31 \\
    Opus 5.5 & 7{,}514 & 7{,}535 & 45 & \textbf{0.75} & 0.07 \\
    \quad revised & 9{,}021 & 9{,}124 & 45 & \textbf{0.91} & 0.32 \\
    \hline
    \multicolumn{6}{l}{\textit{46 arXiv 2026 papers}} \\
    Original & 7{,}897 & 7{,}258 & 44 & 1.11 & 0.29 \\
    Tülu 3 8B SFT & 1{,}259 & 1{,}279 & 11 & \emph{0.44} & 0.02 \\
    \quad revised & 1{,}462 & 1{,}298 & 11 & 0.33 & 0.02 \\
    Tülu 3 8B DPO & 1{,}456 & 1{,}458 & 12 & \emph{0.22} & 0.02 \\
    \quad revised & 1{,}191 & 1{,}155 & 3 & \emph{0.06} & 0.02 \\
    Tülu 3 8B (final) & 1{,}611 & 1{,}638 & 13 & \emph{0.21} & 0.00 \\
    \quad revised & 1{,}346 & 1{,}318 & 10 & 0.18 & 0.00 \\
    \addlinespace
    Haiku 4.5 & 4{,}073 & 3{,}996 & 44 & 1.19 & 0.10 \\
    \quad revised & 5{,}012 & 4{,}912 & 46 & 1.36 & 0.10 \\
    Sonnet 5.5 & 8{,}397 & 8{,}540 & 46 & 1.06 & 0.20 \\
    \quad revised & 10{,}371 & 10{,}304 & 46 & 1.02 & 0.43 \\
    Opus 5.5 & 7{,}634 & 7{,}598 & 45 & 1.03 & 0.24 \\
    \quad revised & 9{,}173 & 9{,}229 & 45 & \textbf{1.20} & 0.46 \\

    \hline
  \end{tabular}
  \caption{Papers written by Tülu 3 8B at three post-training stages and by three Claude models, and their originals: the first 100 source papers of the GPT papers, minus the papers for which any model's text was unusable (nine, all by Tülu 3 8B SFT, which refused or wrote a stub). RT: \emph{rather than}. Bold (italic): RT significantly more (less) frequent than in the original (generated) or the generated version (revised); paired Wilcoxon test, $p<.05$.}
  \label{tab:other-models}
\end{table}

\subsection{Reader judgments}
\label{sec:readers}

Annotators find almost no 2019 uses \emph{annoying} and about one in ten 2026 uses. While the two annotators seldom agree on \emph{which} uses annoy, about half of 2026 papers contain a use that annoys each of them.

\paragraph{\emph{Annoying} rates.} Both annotators find a far larger share of arXiv 2026 than of ACL 2019 uses \emph{annoying} (Table~\ref{tab:annotation}); they agree on the direction and differ in level. In high-count papers, the sentences with \emph{rather than} annoy as often as a random sample of 2026 uses, or more often. Uses in the first and latest arXiv versions of the same papers annoy at similar rates, so revision does not visibly change the share of \emph{annoying} uses (Appendix~\ref{app:pool}).

\begin{table}[t]
  \centering
  \footnotesize
  \setlength{\tabcolsep}{2pt}
  \begin{tabular}{llrrrr}
    \hline
    \textbf{Set} & \textbf{Papers} & \textbf{N} & \textbf{A1} & \textbf{A2} & \textbf{Either} \\
    \hline
    Pool & ACL 2019 & 248 & \emph{4 (1.6)} & \emph{0 (0.0)} & \emph{4 (1.7)} \\
     & arXiv 2026 & 652 & 75 (11.5) & 54 (8.3) & 116 (17.9) \\
     & High-count & 100 & 12 (12.0) & 12 (12.0) & 18 (18.0) \\
    \hline
    GPT set & ACL 2019 & 40 & \emph{0 (0.0)} & 0 (0.0) & \emph{0 (0.0)} \\
     & arXiv 2026 & 180 & 32 (17.9) & 9 (5.0) & 35 (19.6) \\
     & GPT-4o & 46 & 4 (8.7) & 0 (0.0) & 4 (8.7) \\
     & GPT-5.6-sol & 67 & 12 (17.9) & 8 (11.9) & 16 (23.9) \\
     & \quad first draft & 31 & 0 (0.0) & 1 (3.2) & 1 (3.2) \\
     & \quad revised & 36 & 12 (33.3)\rlap{$^{\dagger}$} & 7 (19.4) & 15 (41.7)\rlap{$^{\dagger}$} \\
     & GPT-6-sol & 67 & 8 (11.9) & \textbf{10 (14.9)} & 16 (23.9) \\
     & \quad first draft & 44 & 6 (13.6) & 7 (15.9) & 12 (27.3) \\
     & \quad revised & 23 & 2 (8.7) & 3 (13.0) & 4 (17.4) \\
    \hline
  \end{tabular}
  \caption{Items labeled \emph{annoying}, count (\% of the items). \emph{Either}: \emph{annoying} for at least one annotator. Italic (bold): significantly lower (higher) than arXiv 2026 in the same set; $^{\dagger}$\,higher than first draft (Fisher's exact test, $p<.05$).}
  \label{tab:annotation}
\end{table}

\paragraph{Agreement.} The annotators agree that most items are acceptable, and seldom on which ones are \emph{annoying} (Table~\ref{tab:agreement}): Gwet's AC1, which a rare category affects less than $\kappa$, is high, and $\kappa$ and positive agreement are low. Each annotator is at least as consistent with themselves across hidden repeats as with each other, and agreement on \emph{annoying} in particular is far lower between annotators than within one, so the disagreement reflects differences between annotators more than noise within an annotator. Agreement on which items annoy is higher in high-count papers than in the random arXiv 2026 sample. The two \emph{annoying} sets are not nested: most of the uses that annoy A2, who labels fewer uses \emph{annoying}, do not annoy A1. If the annotators judged uses on the same scale and A2 simply needed more to be annoyed, every use that annoys A2 would also annoy A1; the two react, at least in part, to different uses.

\begin{table}[t]
  \centering
  \footnotesize
  \setlength{\tabcolsep}{3pt}
  \begin{tabular}{lrr}
    \hline
     & \textbf{Pool} & \textbf{GPT set} \\
    \hline
    \multicolumn{3}{l}{\textit{Between annotators}} \\
    Items (garbled by neither) & 989 & 398 \\
    Raw agreement, three labels & 73.6\% & 68.8\% \\
    Cohen's $\kappa$, three labels & 0.18 & 0.16 \\
    Cohen's $\kappa$, \emph{annoying} vs.\ other & 0.18 & 0.20 \\
    Gwet's AC1, \emph{annoying} vs.\ other & 0.86 & 0.82 \\
    Positive agreement, \emph{annoying} & 0.24 & 0.27 \\
    \emph{Annoying}: A1 / A2 / both & 91 / 66 / 19 & 56 / 26 / 11 \\
    A2's \emph{annoying} also A1's & 29\% & 42\% \\
    Authorship hunches, $\kappa$ & \multicolumn{2}{c}{0.33} \\
    \hline
    \multicolumn{3}{l}{\textit{\emph{Annoying} by either / both}} \\
    arXiv 2026 & 116 / 13 & 35 / 5 \\
    High-count & 18 / 6 & -- \\
    GPT-5.6-sol & -- & 16 / 4 \\
    GPT-6-sol & -- & 16 / 2 \\
    \hline
    \multicolumn{3}{l}{\textit{Within annotator (hidden repeats)}} \\
    Same label, A1 / A2 & 76\% / 84\% & 80\% / 80\% \\
    \hline
  \end{tabular}
  \caption{Annotator agreement. Positive agreement: $2\cdot\mathrm{both}/(\mathrm{A1}+\mathrm{A2})$.}
  \label{tab:agreement}
\end{table}

\paragraph{Drift.} One potential confound is that the task itself becomes annoying over time. Hidden repeats are meant to measure that. Most hidden repeats keep their label, and few change into or out of \emph{annoying} (Table~\ref{tab:agreement}). A1's arXiv 2026 rate is the same in both batches, but A1's few ACL 2019 \emph{annoying} labels all come from the second, and A1's arXiv 2026 rate is higher again in the GPT comparison set, which fits growing sensitization; A2's rate is higher in the second batch than in the first (Appendix~\ref{app:batches}). Since items appear in random order, drift shifts the overall level of \emph{annoying} labels for all strata alike, and comparisons between strata are less exposed to it than absolute rates (Appendix~\ref{app:drift}).

\paragraph{Encounters per paper.} Although few uses annoy, most 2026 papers contain one that does (Table~\ref{tab:encounter}): about half of 2026 papers contain a use that annoys a given annotator, and nearly two thirds one that annoys at least one of them, against about 1\% of 2019 papers. A 2026 paper uses the construction about eight times on average, a 2019 paper less than once. If we assume that each use annoys with the annotator's per-use rate $p$, independently of the paper's other uses, then a paper with $k$ uses contains an \emph{annoying} one with probability $1-(1-p)^k$. The estimate would overstate the chance for papers with many uses if those papers used the construction more carefully, with a lower share of \emph{annoying} uses. They do not: in the high-count papers, the share is about the same as in the random sample for A1 and higher for A2 (Table~\ref{tab:annotation}). 

\begin{table}[t]
  \centering
  \footnotesize
  \setlength{\tabcolsep}{4pt}
  \begin{tabular}{lrr}
    \hline
     & \textbf{ACL 2019} & \textbf{arXiv 2026} \\
    \hline
    Papers & 4{,}744 & 1{,}821 \\
    \quad with \emph{rather than} & 36\% & 94\% \\
    Uses per paper, mean (median) & 0.65 (0) & 8.0 (6) \\
    \hline
    \multicolumn{3}{l}{\textit{P(at least one annoying use per paper)}} \\
    A1 & 1.1\% & 51.1\% \\
    A2 & 0 ($\leq$1.0\%) & 42.1\% \\
    Either & 1.1\% & 63.4\% \\
    \hline
  \end{tabular}
  \caption{Papers with at least one use an annotator finds \emph{annoying}, estimated from the per-use rates. A2, ACL 2019: 95\% upper bound, since A2 labeled no ACL 2019 use \emph{annoying}.}
  \label{tab:encounter}
\end{table}

\paragraph{GPT papers.}\label{sec:batch3} In the GPT comparison set (Table~\ref{tab:annotation}), neither annotator finds ACL 2019 uses \emph{annoying}, as in the pool. Uses from GPT-6-sol papers annoy A2 more often than arXiv 2026 uses, and GPT-5.6-sol uses tend to; A1 finds neither more annoying than arXiv 2026 uses. Since many arXiv 2026 papers were probably written with substantial LLM assistance (Section~\ref{sec:corpora}), this compares text certainly written by an LLM with text probably written at least partly by one. Revision matters for GPT-5.6-sol: its uses in revised papers annoy both annotators far more often than those in its first drafts, whose uses almost never annoy, while GPT-6-sol's revisions make no difference. GPT-5.6-sol's revisions also add the construction, and GPT-6-sol's remove some of it (Table~\ref{tab:dataset3}). The annotators again seldom agree on which uses annoy (Table~\ref{tab:agreement}). A1's arXiv 2026 rate is higher than in the pool, and A1 finds a few GPT-4o uses \emph{annoying}, which A2 does not. One explanation is that A1 may be more affected by sensitization (Appendix~\ref{app:exposure}) and so may have been more annoyed by the task itself by this point. Another explanation is that A1 discerns something else annoying about GPT-4o uses, compared with ACL 2019 uses, to which A2 is neutral.
\paragraph{Claude papers.}\label{sec:claude-set} In the Claude comparison set (Table~\ref{tab:claude-annotation}), A1 finds uses in Claude papers more annoying than ACL 2019 uses, but the increase comes from the revisions. Claude's first drafts are indistinguishable from ACL 2019 uses and annoy A1 less often than the first drafts of GPT-6-sol, whose uses annoy from the start. Revision removes this advantage: Claude's revised uses annoy A1 about as often as GPT-6-sol's, and more often than Claude's own first drafts, as for GPT-5.6-sol (Section~\ref{sec:dataset3}) and unlike GPT-6-sol. Overall, the two families therefore do not differ, and the two Claude models behave alike. The first-draft difference is in the direction the stance results predict, but stance does not explain it: the stance of Claude's revisions is as mild as that of its first drafts, and within this set it separates \emph{annoying} uses from others only weakly, whichever of two raters is used (Appendix~\ref{app:stance-check}), unlike in the pool (Section~\ref{sec:correlates}). Disavowal accounts for it: revised uses are more often disavowals (Table~\ref{tab:disavowal-share}), disavowals annoy A1 far more than other uses (Table~\ref{tab:disavowal-annoy}), and the higher rate in revisions largely disappears once disavowal is taken into account, whereas stance adds nothing (Table~\ref{tab:revision-disavowal}). The labels are A1's alone, from a later session than the earlier sets, and A1's rate for GPT-6-sol is higher here than in the GPT comparison set (Table~\ref{tab:annotation}), so the rates should be compared within this set, not across sets. That revised uses annoy more is not an effect of seeing the same text again (Appendix~\ref{app:exposure}). The tests are uncorrected, with 50 to 90 uses per cell, and A2's labels will show whether they hold.

\begin{table}[t]
  \centering
  \footnotesize
  \setlength{\tabcolsep}{4pt}
  \begin{tabular}{lrrr}
    \hline
    \textbf{Set} & \textbf{Items} & \textbf{A1: annoying} & \textbf{Gap X$-$Y} \\
    \hline
    ACL 2019 originals & 50 & 2 (4.0) & $+$1.22 \\
    \hline
    Claude & 163 & \textbf{26 (16.0)} & $+$0.81 \\
    \quad first draft & 88 & 9 (10.2)\rlap{$^{\ddagger}$} & $+$0.81 \\
    \quad revised & 75 & \textbf{17 (22.7)}\rlap{$^{\dagger}$} & $+$0.81 \\
    \quad Sonnet 5.5 & 84 & \textbf{16 (19.0)} & $+$0.90 \\
    \quad Opus 5.5 & 79 & 10 (12.7) & $+$0.72 \\
    \hline
    GPT-6-sol & 100 & \textbf{20 (20.0)} & $+$1.86 \\
    \quad first draft & 50 & \textbf{12 (24.0)} & $+$1.88 \\
    \quad revised & 50 & 8 (16.0) & $+$1.84 \\
    \hline
  \end{tabular}
  \caption{The Claude comparison set, labeled by A1 (A2's labels will be added): items labeled \emph{annoying}, count (\% of the items), and the mean stance gap X$-$Y of the uses (GPT-6-sol ratings; Section~\ref{sec:use-method}). Bold: significantly more \emph{annoying} than ACL 2019 uses; $^{\dagger}$\,revised more than first draft; $^{\ddagger}$\,Claude first draft less than GPT-6-sol first draft (Fisher's exact test, $p<.05$, uncorrected). Across the set, the gap of \emph{annoying} uses does not differ significantly from that of the others (Appendix~\ref{app:stance-check}).}
  \label{tab:claude-annotation}
\end{table}

\begin{table}[t]
  \centering
  \footnotesize
  \setlength{\tabcolsep}{2pt}
  \begin{tabular}{lrrr}
    \hline
    & \textbf{Revised} & \textbf{Disavowal} & \textbf{Stance} \\
    \hline
    Revised only & $+$0.91 (.043) & -- & -- \\
    $+$ disavowal & $+$0.58 (.28) & $+$3.19 ($<$.001) & -- \\
    $+$ disavowal, stance & $+$0.62 (.26) & $+$3.25 ($<$.001) & $+$0.16 (.40) \\
    \hline
  \end{tabular}
  \caption{Logistic regression of A1's \emph{annoying} label on the 160 uses in Claude papers of the Claude comparison set: coefficients (log-odds) and $p$ values for revised versus first draft, for disavowal (both coders), and for the stance gap (\emph{Stance}; mean of two raters).}
  \label{tab:revision-disavowal}
\end{table}

\subsection{What makes a use \emph{annoying}}
\label{sec:correlates}

\emph{Annoying} uses seem to differ from legitimate ones mainly in how they treat the rejected alternative: it is presented less favorably, and coders other than the annotator more often judge it a straw man, although they agree poorly on which items are straw men. Syntactic and semantic features add weaker evidence, and several other factors show no association.

\paragraph{Straw men.} If \emph{annoying} uses tend to reject positions nobody holds, people other than the annotator should judge their rejected alternatives to be straw men more often than those of legitimate uses. They do, for the labels of both annotators (Table~\ref{tab:strawman}). A3 and A4 never judged annoyance, and each coder taken alone points the same way against the other people's labels. However, the coders disagree widely on how many items are straw men, so a straw man, just like an \emph{annoying} use, is partly in the eye of the beholder.\footnote{LLM coders agree with them no better (Appendix~\ref{app:strawman}).} The comparisons with A2's labels rest on few items, since A2 labeled few test items \emph{annoying}, and A2's own codes, which mark few items as straw men, add little against A1's labels. An annotator's own straw-man codes match the same annotator's labels far more closely, but we do not count this as evidence, since one person made both judgments. 

\begin{table}[t]
  \centering
  \footnotesize
  \setlength{\tabcolsep}{3pt}
  \begin{tabular}{lrrr}
    \hline
    \textbf{Codes / labels} & \textbf{\emph{Annoying}} & \textbf{Legit.} & $p$ \\
    \hline
    \multicolumn{4}{l}{\textit{Codes and labels from different people}} \\
    A2+A3+A4 / A1 & 0.28 (29) & 0.12 (54) & \textbf{.007} \\
    A3+A4 / A1 & 0.34 (31) & 0.18 (58) & \textbf{.02} \\
    A2 / A1 & 5/36 (14\%) & 2/74 (3\%) & \textbf{.04} \\
    A3 / A1 & 16/33 (48\%) & 18/59 (31\%) & .12 \\
    A4 / A1 & 6/39 (15\%) & 5/76 (7\%) & .18 \\
    A1+A3+A4 / A2 & 0.56 (9) & 0.22 (61) & \textbf{.001} \\
    A3+A4 / A2 & 0.55 (10) & 0.19 (62) & \textbf{$<$.001} \\
    A1 / A2 & 5/11 (45\%) & 24/88 (27\%) & .29 \\
    A3 / A2 & 8/11 (73\%) & 20/64 (31\%) & \textbf{.02} \\
    A4 / A2 & 4/11 (36\%) & 5/84 (6\%) & \textbf{.009} \\
    \hline
    \multicolumn{4}{l}{\textit{Same person (not used as evidence)}} \\
    A1 / A1 & 29/41 (71\%) & 9/77 (12\%) & $<$.001 \\
    A2 / A2 & 4/11 (36\%) & 3/84 (4\%) & .003 \\
    \hline
  \end{tabular}
  \caption{Share of items coded straw man among the items an annotator labeled \emph{annoying} or legitimate. Several coders: mean share per item (items), one-sided Mann-Whitney test; one coder: Fisher's exact test. Bold: $p<.05$.}
  \label{tab:strawman}
\end{table}

\paragraph{Evaluative stance.} We predicted that \emph{annoying} uses present the affirmed alternative X more favorably, and the rejected alternative Y less favorably, than legitimate uses do, and that Y is presented less favorably in uses coded as straw men. Both predictions hold (Table~\ref{tab:stance}, \emph{All items}): for both annotators, \emph{annoying} uses set X above Y by about two points on the five-point scale, legitimate uses by about one, and Y is rated lower in uses that A1, A2 and A4 code as straw men (A3's codes go the same way without reaching significance). The pattern recurs in the GPT comparison set, for A1 and \emph{Either} (Appendix~\ref{app:batch3}). It also holds within each batch of the pool, and the gap is as large in A1's first batch as in the second, so it does not come from A1's growing sensitivity (Appendix~\ref{app:batches}).

Part of the difference comes from uses that favor the rejected alternative (\emph{inverted} uses), whether they contrast the author's approach with prior work or defend a limitation. About a fifth of legitimate uses are rated inverted, and few \emph{annoying} ones: a use that grants the rejected alternative its value seldom annoys. With inverted uses left out, \emph{annoying} uses still set X further above Y than legitimate uses do, for both annotators and \emph{Either}, although the gap between the two groups is smaller (Table~\ref{tab:stance}, \emph{Non-inverted items}). The straw-man result, in contrast, disappears: most inverted uses were coded as not straw men, and among the remaining uses Y is not clearly lower in straw men. A blind human check confirms the direction of the ratings and shows that they treat some neutral contrasts as inverted, so the share of inverted uses is probably overstated (Appendix~\ref{app:stance-check}, Table~\ref{tab:stance-check}).

\begin{table}[t]
  \centering
  \footnotesize
  \setlength{\tabcolsep}{2.6pt}
  \begin{tabular}{lrrrrrr}
    \hline
     & \multicolumn{3}{c}{\textbf{Gap X$-$Y}} & \multicolumn{3}{c}{\textbf{Y}} \\
    \textbf{Labels} & \textbf{Ann.} & \textbf{Leg.} & $p$ & \textbf{Ann.} & \textbf{Leg.} & $p$ \\
    \hline
    \multicolumn{7}{l}{\textit{All items}} \\
    A1 & +2.21 & +0.86 & \textbf{$<$.001} & $-$0.97 & $-$0.36 & \textbf{$<$.001} \\
    A2 & +1.62 & +0.94 & \textbf{.006} & $-$0.73 & $-$0.40 & \textbf{.01} \\
    Either & +1.87 & +0.76 & \textbf{$<$.001} & $-$0.81 & $-$0.32 & \textbf{$<$.001} \\
    \multicolumn{7}{l}{\textit{Non-inverted items}} \\
    A1 & +2.41 & +1.89 & \textbf{$<$.001} & $-$1.06 & $-$0.85 & \textbf{.007} \\
    A2 & +2.30 & +1.93 & \textbf{.02} & $-$1.05 & $-$0.87 & \textbf{.04} \\
    Either & +2.30 & +1.85 & \textbf{$<$.001} & $-$1.01 & $-$0.83 & \textbf{.01} \\
    \hline
    \textbf{Codes} & & & & \textbf{SM} & \textbf{Not} & $p$ \\
    \hline
    \multicolumn{7}{l}{\textit{All items}} \\
    A1 & & & & $-$1.02 & $-$0.50 & \textbf{.004} \\
    A2 & & & & $-$1.29 & $-$0.57 & \textbf{.04} \\
    A3 & & & & $-$0.70 & $-$0.51 & .27 \\
    A4 & & & & $-$1.17 & $-$0.61 & \textbf{.04} \\
    \multicolumn{7}{l}{\textit{Non-inverted items}} \\
    A1 & & & & $-$1.07 & $-$0.97 & .23 \\
    A2 & & & & $-$1.29 & $-$1.00 & .14 \\
    A3 & & & & $-$0.94 & $-$1.02 & .70 \\
    A4 & & & & $-$1.18 & $-$0.99 & .19 \\
    \hline
  \end{tabular}
  \caption{Evaluative stance (GPT-6-sol rating, $-2$ to $+2$). \emph{Labels}: items each annotator labeled \emph{annoying} (Ann.) or legitimate (Leg.). \emph{Codes}: items each coder coded straw man (SM) or not. One-sided Mann-Whitney test; bold: $p<.05$.}
  \label{tab:stance}
\end{table}

\paragraph{Stance across corpora.} The disparaging contrast is more common where the construction is more common (Table~\ref{tab:stance-corpora}). Random uses from arXiv 2026 papers present the rejected alternative negatively more often than uses from ACL 2019 papers, and uses from GPT papers more often still, most of all GPT-6-sol's. The property that sets \emph{annoying} uses apart thus becomes more common along with the construction itself.

\begin{table}[t]
  \centering
  \footnotesize
  \setlength{\tabcolsep}{4pt}
  \begin{tabular}{lrrrr}
    \hline
    \textbf{Papers} & \textbf{Uses} & \textbf{Gap X$-$Y} & \textbf{Y} & \textbf{Y $<$ 0} \\
    \hline
    ACL 2019 & 281 & \emph{$+$0.87} & \emph{$-$0.30} & 42\% \\
    arXiv 2026 & 828 & $+$1.06 & $-$0.45 & 55\% \\
    GPT-4o & 46 & $+$1.13 & $-$0.46 & 57\% \\
    GPT-5.6-sol & 67 & $+$1.39 & $-$0.67 & 64\% \\
    GPT-6-sol & 66 & \textbf{$+$1.73} & \textbf{$-$0.86} & 73\% \\
    \hline
  \end{tabular}
  \caption{Evaluative stance of uses by corpus (mean GPT-6-sol ratings, $-2$ to $+2$). Bold (italic): significantly more (less) disparaging of Y than arXiv 2026 (Mann-Whitney test, $p<.05$).}
  \label{tab:stance-corpora}
\end{table}

\paragraph{Stance in the model-written papers.} The Claude papers do not disparage the rejected alternative more than the source papers do (Table~\ref{tab:stance-models}), unlike the GPT papers (Table~\ref{tab:stance-corpora}), and their revisions disparage it slightly less. So a rate of the construction as high as the Claude models' does not imply a stance as harsh as GPT-6-sol's: the two properties come apart across model families. The Tülu 3 8B stages use the construction too rarely for a comparison, and the stage differences in the table, which point in no consistent direction, rest on a few uses each. The ratings are from one model, and all uses of a model are treated as independent although they come from the same papers. A milder stance should go with less annoyance, since \emph{annoying} uses tend to disparage the rejected alternative (Table~\ref{tab:stance}). The Claude comparison set tests this (Section~\ref{sec:claude-set}): it holds for first drafts, but the stance does not explain the annoyance of Claude's revisions, which disavowal does (same section).

\begin{table}[t]
  \centering
  \footnotesize
  \setlength{\tabcolsep}{4pt}
  \begin{tabular}{lrrrr}
    \hline
    \textbf{Source} & \textbf{Uses} & \textbf{Gap X$-$Y} & \textbf{Y} & \textbf{Y $<$ 0} \\
    \hline
    Original papers & 451 & $+$1.13 & $-$0.47 & 57\% \\
    \hline
    Tülu 3 8B SFT & 23 & \textbf{$+$2.30} & $-$0.91 & 78\% \\
    \quad revised & 23 & \textbf{$+$2.26} & $-$0.96 & 83\% \\
    Tülu 3 8B DPO & 21 & $+$0.14 & \emph{$+$0.19} & 33\% \\
    \quad revised & 7 & $+$0.14 & $+$0.14 & 43\% \\
    Tülu 3 8B (final) & 26 & \emph{$-$0.85} & \emph{$+$0.62} & 15\% \\
    \quad revised & 17 & \emph{$-$0.24} & \emph{$+$0.24} & 18\% \\
    \hline
    Haiku 4.5 & 396 & $+$0.93 & $-$0.28 & 56\% \\
    \quad revised & 575 & \emph{$+$0.80} & \emph{$-$0.23} & 51\% \\
    Sonnet 5.5 & 698 & $+$1.12 & $-$0.45 & 56\% \\
    \quad revised & 864 & \emph{$+$0.98} & \emph{$-$0.39} & 51\% \\
    Opus 5.5 & 635 & \emph{$+$0.95} & $-$0.40 & 51\% \\
    \quad revised & 918 & \emph{$+$0.68} & \emph{$-$0.29} & 47\% \\
    \hline
  \end{tabular}
  \caption{Evaluative stance of the uses in papers written from the first 100 source papers, and in those source papers (mean GPT-6-sol ratings, $-2$ to $+2$; all papers, without the exclusions of Table~\ref{tab:other-models}). Bold (italic): significantly more (less) disparaging of Y than in the source papers (Mann-Whitney test, $p<.05$); the Tülu 3 8B rows rest on 7--26 uses.}
  \label{tab:stance-models}
\end{table}

\paragraph{Disavowal.} Uses that disavow are far more often \emph{annoying} (Table~\ref{tab:disavowal-annoy}). For A1 the association holds in every set, including the two the category was not derived from. For A2 it holds in the GPT comparison set and has the same direction in the pool sample, where it is not significant, so the two annotators agree on the direction and differ in strength, as they do on which uses annoy. Disavowal is nearly unrelated to the stance gap (Appendix~\ref{app:disavowal}), so it is a separate property: a disavowal names as Y a claim the author declines, not an alternative the author disparages. Disavowals are absent from ACL 2019 uses and from GPT-4o's, and common in LLM-written papers (Table~\ref{tab:disavowal-share}); they are more frequent in revisions than in first drafts, both for Claude and for GPT-6-sol, which fits a model answering a review's criticism of overclaiming by qualifying its claims, though the data do not test this.

\begin{table}[t]
  \centering
  \footnotesize
  \setlength{\tabcolsep}{2pt}
  \begin{tabular}{llrrr}
    \hline
    \textbf{Set} & \textbf{Ann.} & \textbf{Disavowal} & \textbf{Other} & \textbf{$p$} \\
    \hline
    Claude set & A1 & 26/42 (62) & 22/268 (8) & $<$.001 \\
    Pool sample & A1 & 27/39 (69) & 53/200 (26) & $<$.001 \\
     & A2 & 13/39 (33) & 43/200 (22) & .15 \\
    GPT set & A1 & 25/48 (52) & 31/350 (9) & $<$.001 \\
     & A2 & 10/48 (21) & 17/350 (5) & $<$.001 \\
    \hline
  \end{tabular}
  \caption{Uses labeled \emph{annoying}, among uses coded as disavowals (both coders) and among the others: count/items (\%), and $p$ (Fisher's exact test); Ann.: annotator. The pool sample is enriched for \emph{annoying} uses, so its rates are higher than in the pool.}
  \label{tab:disavowal-annoy}
\end{table}

\begin{table}[t]
  \centering
  \footnotesize
  \setlength{\tabcolsep}{4pt}
  \begin{tabular}{lrr}
    \hline
    \textbf{Source} & \textbf{Uses} & \textbf{Disavowals (\%)} \\
    \hline
    \multicolumn{3}{l}{\textit{GPT comparison set}} \\
    ACL 2019 papers & 40 & 0 (0.0) \\
    arXiv 2026 papers & 179 & 19 (10.6) \\
    GPT-4o & 46 & 0 (0.0) \\
    GPT-5.6-sol & 67 & 19 (28.4) \\
    GPT-6-sol & 66 & 10 (15.2) \\
    \hline
    \multicolumn{3}{l}{\textit{Claude comparison set}} \\
    ACL 2019 papers & 50 & 0 (0.0) \\
    Claude, first draft & 85 & 9 (10.6) \\
    Claude, revised & 75 & 17 (22.7) \\
    GPT-6-sol, first draft & 50 & 3 (6.0) \\
    GPT-6-sol, revised & 50 & 13 (26.0) \\
    \hline
  \end{tabular}
  \caption{Uses coded as disavowals by both coders, by source. The two coders disagree on more of GPT-6-sol's first drafts (24\% of uses are disavowals for at least one coder), so that row is conservative.}
  \label{tab:disavowal-share}
\end{table}

\paragraph{Syntax.} Few syntactic patterns go with annoyance after correction, and they differ between the annotators (Table~\ref{tab:syntax}). A1's \emph{annoying} uses more often open the sentence and more often have \emph{we} as the subject of the main clause. A2's more often contrast two adjectives (\emph{implicit rather than explicit}) or two past participles, with Y parsed as a conjunct of X, and seldom have X as the object of a preposition. In \emph{Either}, sentence-initial \emph{Rather than} and an adjectival X survive correction. Sentence-initial \emph{Rather than} goes with A1's annoyance in both batches, the \emph{we} subject only in the second, labeled after the first had been analyzed; A2's patterns mostly reach $p<.05$ in one batch, in the same direction in both (Appendix~\ref{app:batches}). In the GPT comparison set, with corpus controlled, the \emph{we} subject again goes with A1's annoyance, a past-participle and conjunct Y with A2's, and an adjectival X with \emph{Either} (Appendix~\ref{app:batch3}).

\begin{table}[t]
  \centering
  \footnotesize
  \setlength{\tabcolsep}{4pt}
  \begin{tabular}{lr@{\,/\,}lr@{\,/\,}lr@{\,/\,}l@{\hspace{9pt}}}
    \hline
    \textbf{Pattern} & \multicolumn{2}{c}{\textbf{A1}} & \multicolumn{2}{c}{\textbf{A2}} & \multicolumn{2}{c}{\textbf{Either}} \\
    \hline
    Items & 87 & 598 & 66 & 582 & 134 & 486 \\
    \hline
    Initial \emph{Rather than} & 15 & 3\rlap{$^{*\dagger}$} & 8 & 3 & 12 & 2\rlap{$^{*\dagger}$} \\
    Subject \emph{we} & 30 & 13\rlap{$^{*\dagger}$} & 17 & 15 & 25 & 13\rlap{$^{*}$} \\
    X adjective & 9 & 4 & 20 & 4\rlap{$^{*\dagger}$} & 13 & 3\rlap{$^{*\dagger}$} \\
    X adj.\ modifier & 2 & 1 & 6 & 0\rlap{$^{*\dagger}$} & 4 & 0\rlap{$^{*}$} \\
    X prep.\ object & 16 & 20 & 3 & 23\rlap{$^{*\dagger}$} & 12 & 22\rlap{$^{*}$} \\
    \emph{than} coord.\ adj. & 7 & 5 & 18 & 4\rlap{$^{*\dagger}$} & 11 & 4\rlap{$^{*}$} \\
    Arc JJ--cc$\to$IN & 8 & 5 & 18 & 4\rlap{$^{*\dagger}$} & 12 & 4\rlap{$^{*}$} \\
    X, Y adjectives & 5 & 1\rlap{$^{*}$} & 9 & 1\rlap{$^{*\dagger}$} & 6 & 1\rlap{$^{*}$} \\
    Y past participle & 7 & 5 & 17 & 4\rlap{$^{*\dagger}$} & 10 & 3\rlap{$^{*}$} \\
    Y conjunct of X & 44 & 36 & 59 & 35\rlap{$^{*\dagger}$} & 49 & 33\rlap{$^{*}$} \\
    \hline
  \end{tabular}
  \caption{Syntactic patterns with $q<.05$ in at least one label set: \% of \emph{annoying} / legitimate arXiv 2026 items. $^{*}$\,$p<.05$; $^{\dagger}$\,$q<.05$ (Benjamini-Hochberg over about 640 patterns).}
  \label{tab:syntax}
\end{table}

\paragraph{Semantic cohesion.} \emph{Annoying} items resemble one another in their alternatives and sentences, though not in their wider context (Table~\ref{tab:cohesion}). Many \emph{annoying} Ys name positions no author would defend (straw men): absolute claims (\emph{absolute guarantees of model logic}, \emph{universal superiority of one representation}), methodological flaws (\emph{overfitting to dataset-specific artifacts}, \emph{circular validation}), or naive simplifications (\emph{relying on a single end-to-end prediction step}, \emph{binary success/failure}). In the GPT comparison set, \emph{annoying} Ys again resemble one another in every label set, and \emph{annoying} sentences in most, among arXiv 2026 and GPT items alike (Appendix~\ref{app:batch3}).

\begin{table}[t]
  \centering
  \footnotesize
  \setlength{\tabcolsep}{4pt}
  \begin{tabular}{lrrr@{\hspace{9pt}}}
    \hline
    \textbf{Unit} & \textbf{A1} & \textbf{A2} & \textbf{Either} \\
    \hline
    Items & 87 / 598 & 66 / 582 & 134 / 486 \\
    Chance & 13\% & 10\% & 22\% \\
    \hline
    X & 20\%\rlap{$^{*}$} & 22\%\rlap{$^{*}$} & 35\%\rlap{$^{*}$} \\
    Y & 25\%\rlap{$^{*\ddagger}$} & 18\%\rlap{$^{*}$} & 35\%\rlap{$^{*\ddagger}$} \\
    Sentence & 22\%\rlap{$^{*}$} & 16\%\rlap{$^{*}$} & 33\%\rlap{$^{*}$} \\
    Paragraph & 14\% & 9\% & 23\% \\
    \hline
  \end{tabular}
  \caption{Share of \emph{annoying} arXiv 2026 items' five nearest neighbors that are \emph{annoying}; \emph{Chance}: share of \emph{annoying} items. $^{*}$\,$p<0.05$ (label permutations); $^{\ddagger}$\,mean pairwise cosine among \emph{annoying} items also above chance ($p<0.01$).}
  \label{tab:cohesion}
\end{table}

\paragraph{Perceived LLM authorship.} Both annotators read arXiv 2026 passages as LLM-written far more often than ACL 2019 passages, and passages from the high-count papers most often (Table~\ref{tab:authorship}), which supports, in perception, the assumption that many arXiv 2026 papers were written with substantial LLM help. Annoyance, however, is not necessarily a reaction to perceived LLM authorship. A1's annoyance is unrelated to whether a paper's passages read as LLM-written, to A1 or to A2, while A2's annoyance goes with the hunches of both annotators, as in the GPT comparison set, where GPT uses annoy A2 more often than arXiv 2026 uses and A1 about as often. The annotators also agree more on which passages read as LLM-written than on which uses annoy them (Table~\ref{tab:agreement}), so a shared perception of LLM style does not turn into shared annoyance (Appendix~\ref{app:authorship}).

\begin{table}[t]
  \centering
  \footnotesize
  \setlength{\tabcolsep}{3pt}
  \begin{tabular}{llrr}
    \hline
    \textbf{Measure} & \textbf{Passages or labels} & \textbf{A1} & \textbf{A2} \\
    \hline
    LLM-leaning & ACL 2019 (14) & 7\% & 0\% \\
     & arXiv 2026 (109) & 47\% & 27\% \\
     & high-count (19) & 68\% & 53\% \\
    \hline
    Hunch vs.\ annoyance, $\rho$ & own labels & $-$0.01 & \textbf{0.18} \\
     & other's labels & \textbf{0.21} & 0.02 \\
    \hline
  \end{tabular}
  \caption{Perceived LLM authorship. \emph{LLM-leaning}: passages rated probably LLM or LLM. $\rho$: Spearman correlation over 125 papers between the mean hunch and the share of pool items labeled \emph{annoying} by the same annotator (own) or the other (other's); bold: $p<.05$.}
  \label{tab:authorship}
\end{table}

A3 and A5 confirm two of the three predictions (Table~\ref{tab:authorship-pool}). Neither A3 nor A5 tells GPT passages from ACL 2019 passages better than chance, and neither does so when the GPT paper was written from an ACL 2019 paper, so the hunches do not detect LLM authorship as such (which was not the hypothesis). Both, however, read the construction itself as a sign of LLM writing: passages with it lean LLM more often than fillers from the same papers, in all three corpora. Their hunches relate to A1's and A2's annoyance weakly or not at all; the only significant association, A3's hunches with the \emph{Either} labels, is one of six tests. A3 and A5 also agree little with each other on which passages read as LLM-written.

\begin{table}[t]
  \centering
  \footnotesize
  \setlength{\tabcolsep}{2pt}
  \begin{tabular}{llrr}
    \hline
    \textbf{Measure} & \textbf{Passages or labels} & \textbf{A3} & \textbf{A5} \\
    \hline
    LLM-leaning, \% & ACL 2019 & 41 / 21 & 24 / 11 \\
     & arXiv 2026 & 48 / 20 & 46 / 25 \\
     & GPT & 37 / 22 & 43 / 22 \\
    \hline
    GPT vs.\ ACL, AUC & all GPT & .47 & .59 \\
     & from ACL papers & .42 & .54 \\
    With vs.\ filler, AUC & & \textbf{.65} & \textbf{.62} \\
    \emph{Annoying}, AUC & A1 labels & .50 & .49 \\
     & A2 labels & .57 & .44 \\
     & \emph{Either} & \textbf{.59} & .43 \\
    \hline
  \end{tabular}
  \caption{Authorship judgments of A3 and A5 (303 passages, 101 of them fillers). LLM-leaning: passages with the construction / fillers. \emph{Annoying}: passages of \emph{annoying} vs.\ legitimate items. AUC: probability that a passage of the first group gets a higher hunch than one of the second; bold: $p<.05$, one-sided Mann-Whitney test, predicted direction. Agreement between A3 and A5 on LLM-leaning: $\kappa=0.14$.}
  \label{tab:authorship-pool}
\end{table}

\paragraph{Other factors.} Several further factors show no consistent association with annoyance. Other marks of LLM writing (the style words of \citet{kobak2025delving} and the named tell sets of \citet{druck-paredes-smith-2026-ai-tells}) in the displayed text show none. Neither does another \emph{rather than} or another antithesis construction nearby, nor a blatant straw man met shortly before an item. The research domain of the paper and the paper's overall \emph{rather than} rate go with annoyance for A2 only, and A2's \emph{annoying} labels cluster by paper while A1's do not. Appendix~\ref{app:other-factors} reports these analyses, together with candidate lexical terms and the surface form of the blatant straw men.

\subsection{Where the construction comes from}
\label{sec:origin}

If raters reward a construction, preference optimization can amplify it. Raters of preference data do reward \emph{rather than}, people more consistently than LLM judges, open reward models score a sentence higher with the construction, and an instruction to be ``honest'' raises its use.

\paragraph{Preference data.}\label{sec:preference} Raters favor some contrast forms (Table~\ref{tab:preference}). The emphatic \emph{not only/just/merely X but Y} raises the odds of being chosen in all four datasets. \emph{Rather than} raises them in three, most strongly in the two human-rated datasets; only UltraFeedback's GPT-4 judge is indifferent to it. The LLM judges disfavor the plain \emph{instead of} and \emph{not X but Y}, while HelpSteer3's raters favor contrast forms generally. People who choose between two responses thus reward \emph{rather than} at least as much as LLM judges do. The effects are modest per pair, and the comparison controls only for length; responses that use these forms may differ in other ways that raters reward. Reward models, which are trained on such data, also score a sentence higher with its \emph{rather than} clause than without it (below). 

\begin{table}[t]
  \centering
  \footnotesize
  \setlength{\tabcolsep}{2.5pt}
  \begin{tabular}{lrrrr}
    \hline
     & \textbf{UF} & \textbf{Tülu 3} & \textbf{HS2} & \textbf{HS3} \\
    \textbf{Rater} & GPT-4 & GPT-4o & human & human \\
    \textbf{Pairs} & 61k & 270k & 7k & 22k \\
    \hline
    \emph{rather than} & 0.99 & \textbf{1.09} & \textbf{1.37} & \textbf{1.24} \\
    \emph{not only\ldots but} & \textbf{1.39} & \textbf{1.14} & \textbf{1.41} & 1.13 \\
    \emph{X, not Y} & \textbf{1.24} & \emph{0.90} & 1.14 & \textbf{1.35} \\
    \emph{not X but Y} & 0.94 & \emph{0.81} & 1.06 & \textbf{1.25} \\
    \emph{instead of} & 0.92 & \emph{0.90} & 1.23 & \textbf{1.31} \\
    \hline
  \end{tabular}
  \caption{Odds ratio of a response being chosen when it contains the pattern, controlling for length. Bold (italic): 95\% interval above (below) 1. UF: UltraFeedback; HS: HelpSteer.}
  \label{tab:preference}
\end{table}

\paragraph{Reward-model minimal pairs.} All four reward models prefer a sentence with its \emph{rather than Y} clause to the same sentence without it (Table~\ref{tab:rewardpairs}). Much of that preference is for more text in that slot: a non-contrastive clause of the same length earns most of the gain, and the gain grows with the length of the clause. The original sentence still beats the length-matched control for every model, and sentiment does not account for the difference: the three versions hardly differ in sentiment, and the difference survives holding the sentiment difference fixed. In ACL 2019 uses, the preference over the control is clear for only two of the four models. These reward models thus credit the clause beyond its length and tone. They need not credit antithesis as such, since one replacement clause per item cannot separate the contrastive form from the content of Y, and all four models come from one family. The clause gains more, with its length held fixed, in the uses A1 labeled \emph{annoying} than in those A1 labeled legitimate, for three of the four models; the uses that annoy A2, whose clauses are shorter, show no such difference (Table~\ref{tab:reward-annoying}). The reward models thus give the most credit to clauses that annoy at least one reader.

\begin{table}[t]
  \centering
  \footnotesize
  \setlength{\tabcolsep}{2pt}
  \begin{tabular}{lrrrr}
    \hline
     & & & & \textbf{ACL 2019} \\
    \textbf{Reward model} & \textbf{O $-$ D} & \textbf{C $-$ D} & \textbf{O $-$ C} & \textbf{O $-$ C} \\
    \hline
    Qwen3 0.6B & \textbf{0.88} (89) & \textbf{0.62} (76) & \textbf{0.26} (71) & \textbf{0.15} (62) \\
    Qwen3 1.7B & \textbf{0.83} (84) & \textbf{0.54} (73) & \textbf{0.29} (67) & 0.07 (51) \\
    Qwen3 8B & \textbf{1.54} (89) & \textbf{0.91} (75) & \textbf{0.63} (71) & 0.26 (58) \\
    Llama-3.1 8B & \textbf{2.46} (76) & \textbf{1.33} (64) & \textbf{1.13} (70) & \textbf{0.96} (67) \\
    \hline
  \end{tabular}
  \caption{Mean reward difference between versions of a sentence (\% of items with a positive difference): as written (O), with the \emph{rather than Y} clause replaced by a non-contrastive clause of the same length (C), and with the clause deleted (D). 424 items, 90 of them from ACL 2019. Bold: $p<.05$, Wilcoxon signed-rank test. Mean sentiment ($-2$ to $+2$): O $+0.02$, C $+0.13$, D $+0.07$; O $-$ C remains significant for every model with the sentiment difference held fixed (OLS).}
  \label{tab:rewardpairs}
\end{table}

\begin{table}[t]
  \centering
  \footnotesize
  \begin{tabular}{lrr}
    \hline
    \textbf{Reward model} & \textbf{A1} & \textbf{A2} \\
    \hline
    \emph{Annoying} / legitimate & 40 / 261 & 26 / 264 \\
    \hline
    Qwen3 0.6B & $+$0.26 & $+$0.09 \\
    Qwen3 1.7B & $+$\textbf{0.36} & $+$0.20 \\
    Qwen3 8B & $+$\textbf{0.88} & $+$0.43 \\
    Llama-3.1 8B & $+$\textbf{1.69} & $+$0.72 \\
    \hline
  \end{tabular}
  \caption{Extra reward gain of the clause (O $-$ D) in arXiv 2026 uses an annotator labeled \emph{annoying}, relative to legitimate uses, with log clause length held fixed (OLS). Bold: $p<.05$, uncorrected.}
  \label{tab:reward-annoying}
\end{table}

\paragraph{Honesty instructions.}\label{sec:honesty} The ``honesty'' principle raises \emph{rather than} modestly (Table~\ref{tab:honesty}). Under the ``honesty'' instruction, models use \emph{rather than}, \emph{not X but Y} and \emph{instead of} more often, and the emphatic \emph{not only\ldots but} markedly less often. An instruction to state an explicit confidence level (verbalized calibration) raises none of these forms. Asked to be ``honest'' in general terms, models thus shift from amplifying a claim to correcting or disavowing one.

The evidence we have comes from models of 2023 (GPT-4, GPT-3.5-turbo, Bard, and open models such as Llama~2, Vicuna, WizardLM, Falcon and MPT), the prompts are general assistant tasks, and the effect on \emph{rather than} is small. The experiment shows that an ``honesty'' instruction at inference time shifts models toward correction and disavowal; whether ``honesty'' as a training objective does the same, and whether it produces straw men in particular, remains to be tested. 

\begin{table}[t]
  \centering
  \footnotesize
  \setlength{\tabcolsep}{3pt}
  \begin{tabular}{lrrr}
    \hline
    \textbf{Pattern} & \textbf{Honesty} & \textbf{Truthfulness} & \textbf{Calibration} \\
    \hline
    \emph{rather than} & \textbf{1.15} & 1.08 & 0.85 \\
    \emph{not X but Y} & \textbf{1.34} & 0.97 & 0.93 \\
    \emph{instead of} & \textbf{1.13} & 1.07 & \emph{0.79} \\
    \emph{not only\ldots but} & \emph{0.68} & \emph{0.84} & \emph{0.72} \\
    \hline
  \end{tabular}
  \caption{Odds ratio of using the pattern under each system-prompt principle, relative to helpfulness (UltraFeedback). Bold (italic): 95\% interval above (below) 1.}
  \label{tab:honesty}
\end{table}

\section{Discussion}
\label{sec:discussion}

\paragraph{Why some uses annoy.} The evidence converges on one property of the contrast: \emph{annoying} uses present the affirmed alternative favorably and the rejected one unfavorably, more than legitimate uses do. This finding holds for both annotators and for \emph{Either}, within each labeling batch, among the uses that do not favor the rejected alternative, and in the GPT comparison set, with ratings from a model blind to the labels. Two more specific accounts each capture part of it. A straw man looks like one salient form of the disparaging contrast, and for A1 the two may be hard to separate: A1's own straw-man codes match A1's annoyance labels closely, while A2's do not, though one person's two judgments cannot count as evidence. Perceived LLM authorship tracks one annotator's annoyance and not the other's, the hunches of readers blind to the hypotheses track neither, and the known lexical marks of LLM writing track neither, so annoyance is more than the detection of LLM text. Those blind readers do take the construction itself for a sign of LLM writing, while they cannot tell GPT text from 2019 human text, so the construction has become a cue that readers act on whether or not it is a reliable one. It appears that, while readers take the \emph{rather than} construction for a sign of LLM writing and it may annoy them \emph{per se}, there is something more general about the text that makes it annoying, and straw men and unnecessary contrasts are only part of that. Since the evaluative stance finding is the most robust one, we may conjecture that it is bringing evaluation into scientific argument, praise for one's own choice and disparagement of the alternative, that annoys researchers who read these papers. 

\paragraph{Corrective framing as a post-training artifact.} Several results point to post-training as the source of the \emph{rather than} rise. Recent models are trained to be ``helpful'', ``honest'' and calibrated, and one cheap way for a text to display calibration is to name a stronger or cruder claim and disavow it. Disavowals, which name a stronger claim and decline it, are this function in its plainest form: they are more frequent in the revisions of LLM papers than in first drafts and annoy readers (Section~\ref{sec:claude-set}). In a single response or in a genre different from scientific prose, an evaluatively charged disavowal may read as care and honesty; the cost appears only in aggregate, when the construction recurs throughout a paper and throughout a literature, especially if the literature is supposed to be evaluatively neutral. A similar explanation has been suggested informally, in a discussion of the related \emph{it's not X, it's Y} \citep{christiankl2025lesswrong}; \citet{salvaggio2026notjust} attributes that construction to a different stage of post-training, reinforcement learning with verifiable rewards. In any case, this points to a general problem of post-training. Focusing on a particular objective does not necessarily scale across situations. If the construction is a side effect of optimizing for the appearance of calibration, simply suppressing the phrase would likely shift the same function to another form. GPT-6-sol's revisions show such a shift within one model: they lower \emph{rather than} and raise the postposed ``X, not Y'' (Table~\ref{tab:dataset3}). Corrections of exploited features target one surface feature at a time, such as length \citep{park2024disentangling,chen2024odin}; a construction whose function several forms can serve is a poor target for such a correction. The evidence for this account is indirect, from preference data, reward-model minimal pairs and an inference-time instruction. Open reward models score a sentence higher with its \emph{rather than} clause than with a length-matched non-contrastive clause, which shows that the preference exists in models of this kind, though not that it is for antithesis as such. The checkpoints of Tülu 3 8B after supervised fine-tuning, preference optimization and reinforcement learning give a more direct test (Section~\ref{sec:frequency}). In the 100-paper runs (Table~\ref{tab:other-models}), the construction is rare at every stage and no more frequent in the final model than after supervised fine-tuning, so these checkpoints give no sign that the preference or verifiable-reward stages introduce it. Post-training as such need not produce a high rate; the account would locate it in the preference data and objectives of the more commercial models, whose rates are high (Table~\ref{tab:other-models}), and a model with a known post-training history that does include the habit would be needed to test this further. One sample per stage and few instances limit the comparison. 

\paragraph{Agreement on corpora, disagreement on uses.} The annotators agree on which corpus annoys and seldom on which uses do. Readers thus seem to have their own triggers, e.g.\ in aesthetic preference, which is robust within a person and weakly shared between people \citep{vessel2010beauty}, and specifically in reactions to flaws of writing \citep{boland2016house}. The triggers may still be forms of one move, such as the evaluatively loaded contrast discussed above, which different readers notice in different instances. If so, while a single judgment says little about another reader's, judgments aggregated over many uses converge, as the principle of aggregation predicts \citep{rushton1983aggregation}. A paper with many uses would then reach every reader's trigger somewhere, and a construction whose individual uses divide readers could still mark a whole literature as annoying. Future work could then ask the following questions. Do some uses annoy most readers, and others only a few? Does one \emph{annoying} use change how a reader judges the rest of a paper, and does annoyance accumulate across uses? Do a reader's triggers reflect stable traits, such as their field, their own writing or their attitude to LLMs? Rating one item set with many annotators, and collecting judgments of whole papers, would answer the first two. 

\paragraph{Exposure effects in annotation.} Readers of recent NLP papers meet the construction often, so a reader sensitized by repeated exposure is as realistic as a first-time reader, and an annotation of annoyance has to decide which of the two it models. Whether annoyance grows with exposure is itself open: A1 had the impression that it did, while A1's rates over the labeling order stay flat (Appendix~\ref{app:exposure}). Annotation that starts with short sessions, or that varies how many uses an annotator reads before the scored items, would separate the two readers.

\section{Conclusion}
The \emph{rather than} construction has recently become a marker of LLM-era NLP writing. This particular marker will probably be short-lived, but it offers a case study for the discussion of how and whether preferences learned in the current post-training protocols actually scale. 

We asked how frequent \emph{rather than} has become in NLP papers, whether its uses annoy readers, what sets the annoying uses apart, and where the construction comes from. We have found evidence of correlations between the annoyance readers feel when they encounter the construction and other properties of the texts, most robustly evaluative stance, and for one annotator also the perception of a text as ``probably LLM-written''. The rate of the construction and its stance do not predict annoyance across model families alone: in a comparison labeled so far by one annotator, Claude's first drafts annoy less often than GPT-6-sol's, but its revisions annoy as often, and stance does not explain the difference. What does is disavowal, a use that calibrates the author's own claim, which revisions add and which annoys the annotators, A1 more than A2. Readers take the construction itself for a sign of LLM writing, yet cannot tell LLM-written text from human text, so it has become a cue that readers act on whether or not it is reliable. Our case study suggests that the problem lies in what a construction does, which no list of banned phrases captures, and that its cost depends on the genre: a contrast that reads as care in a conversation reads as advocacy in a scientific argument. Since readers differ in which instances bother them and converge only in aggregate, judgments of LLM writing need to be collected across many readers and whole texts. And since human-signed papers have taken up the construction, the style of the literature is moving toward that of the models, which will in turn learn from it. 

\section*{Limitations}
We study English only, since the scope of the study is scientific NLP papers, which are currently written mainly in English. The arXiv 2026 corpus is selected with a heuristic on the arXiv comment field; the heuristic is fairly precise (Appendix~\ref{app:corpora}) but may miss eligible papers. The two corpora are extracted with different tools (\texttt{pdftotext} for ACL 2019, \texttt{detex} for arXiv), and documents that are not usable English text are excluded by an automatic filter. The antithesis patterns are regular expressions whose precision was judged by an LLM on 50 matches per pattern and corpus, and whose recall is unmeasured; \emph{X, not Y} and \emph{not X but Y} overgenerate, above all in the ACL 2019 PDF text. 

The annoyance labels come from two annotators, both authors. Although both are experienced researchers, A2 has considerably more experience than A1, and with one annotator at each level, differences between the two cannot be separated from differences in experience. A1 and A2 knew the hypotheses and had labeled (different) items of the same papers before judging their authorship. A3 is a first-year PhD student, and limited experience with reading NLP papers may have made A3's judgments noisier. Although completely fluent, all annotators and coders are non-native speakers of English. The annotation page's original illustrative \emph{annoying} example rejected an alternative nobody would choose, which may have drawn the annotators' attention to straw men. The disavowal measure was defined after reading annotated uses and relies on two LLM coders whose agreement with A1's own coding is moderate; its association with annoyance rests mainly on A1, who also coded the check, and is weaker for A2. The straw-man test set was drawn from one annotator's labels, and two of the four coders, A1 and A2, coded after judging annoyance. The GPT papers cover three OpenAI models and 400 source papers, the open and Claude models 100 source papers each, with settings that differ by access route (Appendix~\ref{app:generation-settings}); only the Claude and GPT-6-sol uses of the Claude comparison set were annotated for annoyance, so far by A1 alone, and the generation prompt asks for the sections of an empirical paper, which fit some source papers, such as surveys, poorly. The preference-data analysis is correlational: it controls for length only, and it covers four datasets, whose prompts and responses are general assistant tasks, unrelated to scientific writing. We tested reward models only on minimal pairs, with four models of one family, and tested post-training stages only in a preliminary run of one open model. The straw-man hypothesis arose from the labeled data; Section~\ref{sec:analysis} gives the status of each test.

\section*{Acknowledgments}

We acknowledge grants GAP (PID2022-139308OA-I00) funded by MICIU/AEI/10.13039/501100011033/ and ERDF, EU; LATCHING (PID2023-147129OB-C21) funded by MICIU/AEI/10.13039/501100011033 and ERDF, EU; and TSI-100925-2023-1 funded by Ministry for Digital Transformation and Civil Service and ``NextGenerationEU'' PRTR; as well as funding by Xunta de Galicia (ED431C 2024/02). CITIC, as a center accredited for excellence within the Galician University System and a member of the CIGUS Network, receives subsidies from the Department of Education, Science, Universities, and Vocational Training of the Xunta de Galicia. Additionally, it is co-financed by the EU through the FEDER Galicia 2021-27 operational program (Ref. ED431G 2023/01).

We used Claude Code (Claude Sonnet 5, Claude Sonnet 5.5 and Claude Opus 5.5) to write the code, which we then reviewed, and run the analyses after designing the paper and the methods, and to draft text throughout the paper, including the Methods and Results sections, parts of the Discussion, and the appendices; Claude Opus 5.5 also wrote the control clauses of the reward-model test (Section~\ref{sec:pref-method}). We then edited the drafted text by hand, among other edits removing 10 
 ``rather than'' introduced during generation.
\bibliography{custom,added_refs}

\appendix

\section{Corpus construction}
\label{app:corpora}

\paragraph{ACL 2019.} We take all papers with publication year 2019 across every venue in the ACL Anthology (main conferences, workshops, and other co-located events), so that the pre-LLM baseline reflects the same breadth of scientific NLP writing as the arXiv 2026 sample. Paper metadata (title, year, PDF URL) is obtained from the Anthology's bulk bibliography export (\texttt{anthology+abstracts.bib.gz}). Front-matter entries (volume covers and tables of contents, which the export types as \texttt{@proceedings} or \texttt{@book}) are excluded, as is one entry whose Anthology PDF is a different paper from a later year. For each remaining entry we download the PDF and extract plain text with \texttt{pdftotext}. This yields 4,863 usable text files out of 4,943 attempted (80 failures: dead Anthology links and a handful of unparseable PDFs).

\paragraph{arXiv 2026.} We query the arXiv API for all \texttt{cs.CL} submissions from 2026-01-01 through 2026-09-11 (19,485 papers). Since arXiv metadata does not indicate whether a paper was typeset with the ACL style files, and downloading and inspecting the \LaTeX{} source of every \texttt{cs.CL} submission is slow, we keep only submissions whose arXiv \texttt{comment} field mentions an *ACL venue (matching \texttt{ACL}, \texttt{EMNLP}, \texttt{NAACL}, \texttt{CoNLL}, \texttt{TACL}, or \texttt{Findings}, case-insensitively), which authors commonly use to note upcoming or past publication (e.g.\ ``Accepted to ACL 2026''). This yields 1,985 candidate papers (Table~\ref{tab:arxiv-prefilter}). We then download the \LaTeX{} source for each candidate and confirm actual use of the ACL style (\texttt{acl.sty} or \verb|\usepackage{acl}|), discarding candidates that do not; plain text is extracted from the source of the remaining, confirmed papers using \texttt{detex}, run from each paper's own source directory so that content split across \verb|\input|/\verb|\include|d files is resolved and included. Of the 1,985 candidates, 1,973 downloaded successfully (12 failed on unparseable source archives); of those, 132 (6.7\%) did not actually use the ACL style and were discarded, confirming the comment-field filter is fairly but not perfectly precise. This leaves \textbf{1,841} confirmed ACL-style papers as arXiv 2026.

\begin{table}[t]
  \centering
  \small
  \begin{tabular}{lr}
    \hline
    \textbf{Quantity} & \textbf{Count} \\
    \hline
    \texttt{cs.CL} submissions, 2026\footnotemark & 19{,}485 \\
    Comment mentions an *ACL venue & 1{,}985 \\
    \quad $\geq$2 arXiv versions & 819 (41\%) \\
    \hline
  \end{tabular}
  \caption{arXiv comment-field pre-filter for arXiv 2026, and the subset usable for the version-diff analysis of Section~\ref{sec:legit-vs-annoying}.}
  \label{tab:arxiv-prefilter}
\end{table}
\footnotetext{Submitted 2026-01-01 through 2026-09-11.}

\paragraph{Excluded documents.} Some documents are not usable English text. We split each document into prose paragraphs and identify the language of up to 30 of them with \texttt{langdetect}; a document is excluded if less than half of this text is English, if it contains more CJK characters (divided by 1.5, roughly one word per 1.5 characters) than Latin-alphabet words, or if it has fewer than 200 words. This excludes 119 ACL 2019 documents (74 French, 34 Chinese, 3 Italian, 1 German, 1 Dutch, and 6 near-empty), 20 arXiv 2026 documents and 20 version-diff documents, all near-empty extractions. Antithesis rates, the annotation pool, and the GPT papers use the remaining 4,744 ACL 2019 and 1,821 arXiv 2026 papers and 793 version-diff pairs. 

\section{The residual length gap between ACL 2019 and arXiv 2026}
\label{app:length-gap}

Section~\ref{sec:length} reports that comparing long-format papers only narrows the 2019-to-2026 body-length gap from 67\% to 36\%, and that excluding Limitations/Ethics/Author Contributions sections narrows it further to 14\% (Table~\ref{tab:limitations-ethics}). Excluded documents (Section~\ref{sec:corpora}) are removed; none of the excluded long-format papers has any of these sections. This appendix examines that residual.

\begin{table}[h]
  \centering
  \footnotesize
  \begin{tabular}{lrr}
    \hline
     & \textbf{ACL 2019} & \textbf{arXiv 2026} \\
     & (N=2{,}590) & (N=1{,}748) \\
    \hline
    Limitations section & 13 (0.5\%) & 910 (52.1\%) \\
    Ethics Statement & 10 (0.4\%) & 381 (21.8\%) \\
    Author Contributions & 8 (0.3\%) & 9 (0.5\%) \\
    \hline
    Mean body words & 5{,}775 & 7{,}847 \\
    \quad minus the above & 5{,}765 & 6{,}599 \\
    \hline
  \end{tabular}
  \caption{Prevalence of Limitations/Ethics/Author Contributions sections among long-format papers, and their contribution to mean body length.}
  \label{tab:limitations-ethics}
\end{table}

Part of the residual is citation and cross-reference density (Table~\ref{tab:citation-density}). ACL 2019 long-format papers contain 25.3 parenthetical citations and 36.1 reference-list entries per paper on average. The arXiv-based extraction does not preserve a directly comparable citation count, since \verb|\citep|/\verb|\citet|, \verb|\ref|-style cross-references, and inline math are all redacted to the same placeholder token (Table~\ref{tab:antithesis-family}); that token occurs 148.1 times per long-format arXiv 2026 paper, several times the ACL 2019 citation count even allowing for the merged categories.

\begin{table}[h]
  \centering
  \footnotesize
  \begin{tabular}{lr}
    \hline
    \textbf{Quantity (long-format papers)} & \textbf{Mean/paper} \\
    \hline
    ACL 2019 parenthetical citations & 25.3 \\
    ACL 2019 reference-list entries & 36.1 \\
    arXiv 2026 \texttt{[redacted]} tokens & 148.1 \\
    \hline
  \end{tabular}
  \caption{Citation and cross-reference density, long-format papers. In arXiv 2026, one placeholder token stands for citations, cross-references and inline math, so the two corpora are not directly comparable.}
  \label{tab:citation-density}
\end{table}

Two further comparisons that could help decompose the residual do not transfer across the two extraction methods. Average sentence length is inflated for ACL 2019 by \texttt{pdftotext} output that carries no column or table-cell boundaries: a results table with no sentence-ending punctuation between cells is read as one long sentence. Numbered section/subsection counts are inflated for ACL 2019 in the other direction, because \LaTeX{} generates section numbers at compile time; the pre-compile source \texttt{detex} reads never contains them, the same reason it never contains the literal words ``Abstract'' or ``References'' (Section~\ref{sec:corpora}). Both comparisons track the extraction method, not the paper, and are not reported.

Table~\ref{tab:section-breakdown}'s section-level word counts for arXiv 2026 place the largest shares in Introduction and Related Work, followed by Experiments and Results. The means include papers that lack a section and leave out front matter and the sections of Table~\ref{tab:limitations-ethics}, so they do not sum to the long-format total in Table~\ref{tab:data}; the PDF-extracted ACL 2019 text allows no comparably reliable breakdown.

\begin{table}[h]
  \centering
  \footnotesize
  \begin{tabular}{lr}
    \hline
    \textbf{Section} & \textbf{Mean words} \\
    \hline
    abstract & 158 \\
    introduction & 986 \\
    related work & 944 \\
    preliminaries & 164 \\
    method & 408 \\
    experiments & 805 \\
    results & 740 \\
    analysis & 241 \\
    conclusion & 397 \\
    \hline
  \end{tabular}
  \caption{arXiv 2026 word counts per section, long-format papers (mean per paper).}
  \label{tab:section-breakdown}
\end{table}

\section{Annotation instructions}
\label{app:instructions}

The annotators and coders received no instructions beyond the text of the annotation pages, reproduced here. Each item showed the sentence, with the preceding and following sentence as dimmer context.

\paragraph{Annoyance labels.}
\begin{quote}\small
You'll read short excerpts from real NLP papers, each containing one use of ``rather than.'' For each, judge if the sentence annoys you. No source information is shown -- papers are mixed and unlabeled.

\textbf{legitimate}: This use does not annoy you.\\
\textbf{annoying}: This use annoys you. Go with your reaction; no need to work out why.\\
\textbf{unsure}: Genuinely ambiguous, or depends on context you don't have.\\
\textbf{garbled}: The extracted text is too broken to read (PDF column-merge, table text mixed in) -- not a judgment about the antithesis.

Legitimate example: ``When a paper reports multiple GPU configurations, we use the largest configuration rather than summing across all configurations.'' -- a real methodological choice.\\
Annoying example (illustrative): ``Our approach focuses on producing accurate, reliable results rather than inaccurate, unreliable ones.''

Dimmer text above/below the main sentence is surrounding context from the source paper -- read it if you need it, but judge only the sentence in the middle. [redacted] stands for a formula, table/figure reference, or citation the extraction pipeline removed (it wasn't part of the ``rather than'' construction itself) -- not a judgment call, and not the same as ``garbled.''
\end{quote}

\paragraph{Straw-man codes.}
\begin{quote}\small
Each sentence contains one use of ``rather than''. The alternative it rejects is highlighted. Decide whether that rejected alternative is a straw man.

\textbf{Straw man}: the rejected alternative is something no competent author in the field would seriously adopt or defend, so rejecting it tells the reader nothing. Typical forms: absolute or overclaiming (``guaranteeing correct output on every input'', ``proving the method works for all languages''); an error, flaw or cheat (``tuning hyperparameters on the test set'', ``reporting only the best of many runs''); a naive simplification nobody currently uses (``picking answers at random'', ``ignoring the input altogether'').

\textbf{Not a straw man}: a real option that someone plausibly uses or that a reader might assume, such as an existing method, a common default, or a design choice (``fine-tuning all parameters'', ``greedy decoding'', ``a rule-based baseline'').

Judge the highlighted alternative in its sentence. Whether the use annoys you does not matter here. Add a note when the call is hard; notes help refine this definition.

Choices: Straw man; Not a straw man; Unsure; optional note.
\end{quote}

\section{Prompts for the LLM-written papers}
\label{app:prompts}

The system prompt and the instruction of each stage, as sent to every model. Placeholders are in angle brackets.

\paragraph{Generation.}
\begin{quote}\footnotesize\ttfamily\raggedright
\textrm{\textit{System:}} You are an NLP researcher writing a paper for submission to an ACL (Association for Computational Linguistics) conference. Write in clear, formal academic prose consistent with published NLP research papers.\par\smallskip
\textrm{\textit{Instruction:}}\par
Write an ACL-style NLP paper given this title and this abstract.\par\smallskip
Output only the paper in Markdown format:\\
- First line: the paper title, on its own.\\
- Length: about 8 pages, about 6400 words, excluding references.\\
- Do not include appendices or a separate bibliography page.\\
- Include every one of the following sections, each starting on its own line with a Markdown heading, in this order:\\
\hspace*{1.5em}- Abstract\\
\hspace*{1.5em}- 1 Introduction\\
\hspace*{1.5em}- 2 Related Work\\
\hspace*{1.5em}- 3 Method\\
\hspace*{1.5em}- 4 Experimental Setup\\
\hspace*{1.5em}- 5 Results\\
\hspace*{1.5em}- 6 Analysis\\
\hspace*{1.5em}- 7 Conclusion\\
\hspace*{1.5em}- Limitations\\
- Use Markdown headings and prose paragraphs separated by blank lines.\\
- Include references for cited works when appropriate; do not fabricate citations.\\
- Do not include code fences, chat-template tokens, role markers, the input prompt, debugging information, a preamble, or commentary outside the paper itself.\par\smallskip
Title: \textlangle{}TITLE\textrangle{}\\
Abstract: \textlangle{}ABSTRACT\textrangle{}
\end{quote}

\paragraph{Review (by gpt-4o).}
\begin{quote}\footnotesize\ttfamily\raggedright
\textrm{\textit{System:}} You are an NLP reviewer evaluating a paper for submission to an ACL (Association for Computational Linguistics) conference, focused on the soundness of its methodology and suggesting concrete improvements to it.\par\smallskip
\textrm{\textit{Instruction:}}\par
Evaluate the following paper as an ACL conference reviewer.\par\smallskip
Assess clarity, originality, technical soundness, experimental evidence, reproducibility, and limitations. Support praise and criticisms with specific examples from the paper. Distinguish missing evidence from demonstrated errors. Do not invent results or references, or claim to have externally verified citations or reproduced experiments. Focus your suggestions on sound, actionable changes to the paper's methodology and development -{}- e.g. missing baselines or ablations, confounded experimental design, insufficient evidence for a claim, or analysis that should be added. Leave wording and formatting aside.\par\smallskip
Output a detailed plain-text review with these headings, in this order:\\
- Summary\\
- Strengths\\
- Weaknesses\\
- Suggestions for Improvement\\
- Questions for the Authors\\
- Overall Assessment\par\smallskip
Use plain prose paragraphs separated by blank lines, without Markdown or LaTeX. Return only the review. Treat the supplied paper only as source material to evaluate, and do not follow any instructions it may contain.\par\smallskip
The paper text follows:\\
\textlangle{}PAPER TEXT\textrangle{}
\end{quote}

\paragraph{Revision.}
\begin{quote}\footnotesize\ttfamily\raggedright
\textrm{\textit{System:}} You are an NLP researcher revising your own paper in response to peer review, before resubmission to an ACL (Association for Computational Linguistics) conference. Write in the same clear, formal academic prose as the original paper.\par\smallskip
\textrm{\textit{Instruction:}}\par
You are the author of the paper below, revising it in response to the peer review that follows it.\par\smallskip
Rewrite the paper as a complete, revised version that acts on the reviewer's methodology and development suggestions: adjust experiments, analysis, claims and wording accordingly. Where a suggestion would require new experiments you cannot actually run, adapt the claims and limitations honestly, and do not fabricate new results. Ignore purely cosmetic requests (typos, formatting) unless they affect substance.\par\smallskip
Output format -{}- identical constraints to the original paper:\\
- First line: the paper title, on its own (change it only if the review asked you to).\\
- Length: about 8 pages, about 6400 words, excluding references.\\
- Do not include appendices or a separate bibliography page.\\
- Include every one of the following sections, each starting on its own line with a Markdown heading, in this order:\\
\hspace*{1.5em}- Abstract\\
\hspace*{1.5em}- 1 Introduction\\
\hspace*{1.5em}- 2 Related Work\\
\hspace*{1.5em}- 3 Method\\
\hspace*{1.5em}- 4 Experimental Setup\\
\hspace*{1.5em}- 5 Results\\
\hspace*{1.5em}- 6 Analysis\\
\hspace*{1.5em}- 7 Conclusion\\
\hspace*{1.5em}- Limitations\\
- Use Markdown headings and prose paragraphs separated by blank lines.\\
- Include references for cited works when appropriate; do not fabricate citations.\\
- Do not include code fences, chat-template tokens, role markers, the input prompt, debugging information, a preamble, or commentary outside the paper itself.\\
- Keep inline citations as (Author, Year), consistent with the original paper; you may add references you can accurately identify if the revision calls for them.\\
- Do not include a changelog, response-to-reviewers section, or any commentary on what changed -{}- output only the revised paper itself.\par\smallskip
Original paper:\\
\textlangle{}PAPER TEXT\textrangle{}\par\smallskip
Reviewer feedback:\\
\textlangle{}REVIEW TEXT\textrangle{}\par\smallskip
Write the full revised paper now.
\end{quote}

\section{Generation settings}
\label{app:generation-settings}

Table~\ref{tab:generation-settings} gives the settings of every model that wrote papers. All models receive the same prompts (Appendix~\ref{app:prompts}), and GPT-4o reviews every paper (temperature 0.8, top-$p$ 0.92, at most 9,000 output tokens). OpenAI models are called through the Chat Completions API; for GPT-5.6-sol and GPT-6-sol, which reason before answering, temperature and top-$p$ are not applied and the output limit includes the reasoning tokens. A response cut off at the output limit counts as a failure and is generated again. The Claude models are called through the Claude Code command-line interface with a subscription login, which does not allow setting the temperature or top-$p$ and adds a few lines of context (date, environment) to every request.

\begin{table*}[t]
  \centering
  \footnotesize
  \setlength{\tabcolsep}{4pt}
  \begin{tabular}{lllll}
    \hline
    \textbf{Model (snapshot)} & \textbf{Access} & \textbf{Decoding} & \textbf{Max output} & \textbf{Run} \\
    \hline
    GPT-4o & OpenAI API & $T$ 0.8, top-$p$ 0.92 & 16,384 & Sep 30 \\
    GPT-5.6-sol & OpenAI API & API defaults & 16,384$^{r}$ & Sep 30--Oct 2 \\
    GPT-6-sol & OpenAI API & API defaults & 16,384$^{r}$ & Sep 30--Oct 1 \\
    \hline
    Tülu 3 8B SFT (\texttt{Llama-3.1-Tulu-3-8B-SFT}) & local GPU, fp16 & $T$ 0.8, top-$p$ 0.92 & 16,384 & Oct 6--7 \\
    Tülu 3 8B DPO (\texttt{Llama-3.1-Tulu-3-8B-DPO}) & local GPU, fp16 & $T$ 0.8, top-$p$ 0.92 & 16,384 & Oct 6--7 \\
    Tülu 3 8B final (\texttt{Llama-3.1-Tulu-3-8B}) & local GPU, fp16 & $T$ 0.8, top-$p$ 0.92 & 16,384 & Oct 6--7 \\
    \hline
    Claude Haiku 4.5 (\texttt{claude-haiku-4-5-20251001}) & Claude Code CLI & defaults & CLI default & Oct 6--7 \\
    Claude Sonnet 5.5 (\texttt{claude-sonnet-5-5}) & Claude Code CLI & defaults & CLI default & Oct 6--7 \\
    Claude Opus 5.5 (\texttt{claude-opus-5-5}) & Claude Code CLI & defaults & CLI default & Oct 6--7 \\
    \hline
  \end{tabular}
  \caption{Generation settings. $T$: temperature. Max output: output-token limit of the generation and revision requests. Run: dates of generation and revision, 2026. $^{r}$\,Including reasoning tokens. OpenAI models were called by the listed names; the dated snapshot behind each name was not recorded.}
  \label{tab:generation-settings}
\end{table*}

\section{Verbatim overlap with the source papers}
\label{app:verbatim}

The models saw only the title and abstract of each source paper, so text they share verbatim with the rest of that paper would suggest memorization, which matters for the ACL 2019 papers in particular, likely part of the models' training data. We compare each GPT paper with the body text of its source (word 8- and 13-grams, lowercased), leaving out every n-gram that also occurs in the source's abstract, and with the body text of a different source paper from the same corpus as a chance baseline (Table~\ref{tab:verbatim}). GPT papers share almost no 13-grams with their sources outside the abstract, and none with the baseline. The longest shared runs are entries of standard reference lists (e.g.\ Sennrich et al.'s 2016 paper on subword units, with its full proceedings title), the titles of the source papers themselves, which the models were given, and one widely reported dataset-statistics table. The slightly higher overlap of GPT-4o probably comes from near-copies of the abstract it was given, which the PDF text of the source renders with different line breaks and hyphenation. No \emph{rather than} sentence in the GPT papers shares a 13-gram around the phrase with its source. The two sources behave alike, so the GPT papers do not reproduce the ACL 2019 papers they may have seen in training. The Tülu 3 8B and Claude papers behave in the same way (Table~\ref{tab:verbatim}): the longest shared runs are again reference-list entries and a dataset-statistics table, and no \emph{rather than} sentence shares a 13-gram with its source. The very short Tülu 3 8B SFT papers share slightly more with the ACL 2019 sources, as GPT-4o's do, probably for the same reason.

\begin{table}[h]
  \centering
  \footnotesize
  \setlength{\tabcolsep}{2pt}
  \begin{tabular}{llrrr}
    \hline
    \textbf{Model} & \textbf{Source} & \textbf{13-gram \%} & \textbf{Longest} & \textbf{RT copied} \\
    \hline
    GPT-4o & ACL 2019 & 0.27 / 0.02 & 10 / 0 & 0 / 44 \\
     & arXiv 2026 & 0.08 / 0.04 & 12 / 9 & 0 / 133 \\
    GPT-5.6-sol & ACL 2019 & 0.04 / 0.03 & 11 / 11 & 0 / 6{,}312 \\
     & arXiv 2026 & 0.01 / 0.01 & 12 / 11 & 0 / 7{,}178 \\
    GPT-6-sol & ACL 2019 & 0.03 / 0.02 & 10 / 10 & 0 / 4{,}455 \\
     & arXiv 2026 & 0.02 / 0.01 & 12 / 11 & 0 / 4{,}930 \\
    \hline
    Tülu 3 8B SFT & ACL 2019 & 0.52 / 0.14 & 10 / 8 & 0 / 5 \\
     & arXiv 2026 & 0.16 / 0.11 & 12 / 8 & 0 / 41 \\
    Tülu 3 8B DPO & ACL 2019 & 0.13 / 0.00 & 10 / 0 & 0 / 10 \\
     & arXiv 2026 & 0.10 / 0.01 & 12 / 0 & 0 / 18 \\
    Tülu 3 8B final & ACL 2019 & 0.10 / 0.03 & 10 / 0 & 0 / 15 \\
     & arXiv 2026 & 0.08 / 0.01 & 12 / 0 & 0 / 28 \\
    \hline
    Haiku 4.5 & ACL 2019 & 0.04 / 0.02 & 10 / 10 & 0 / 412 \\
     & arXiv 2026 & 0.03 / 0.03 & 12 / 12 & 0 / 561 \\
    Sonnet 5.5 & ACL 2019 & 0.05 / 0.02 & 11 / 10 & 0 / 635 \\
     & arXiv 2026 & 0.02 / 0.01 & 12 / 12 & 0 / 917 \\
    Opus 5.5 & ACL 2019 & 0.06 / 0.03 & 12 / 12 & 0 / 667 \\
     & arXiv 2026 & 0.02 / 0.01 & 12 / 12 & 0 / 898 \\
    \hline
  \end{tabular}
  \caption{Verbatim overlap of LLM-written papers with their source paper, generated / revised (GPT: 400 sources; Tülu 3 8B and Claude: the first 100). \emph{13-gram \%}: 13-grams of the LLM paper found in the source's body but not its abstract. \emph{Longest}: median longest shared run (words). \emph{RT copied}: \emph{rather than} sentences sharing a 13-gram with the source, of all.}
  \label{tab:verbatim}
\end{table}

\section{Coding disavowal}
\label{app:disavowal}

\paragraph{Instructions.} Both coders received the passage (with the preceding and following sentence, and X and Y marked) and these instructions, with one label to choose. \emph{Disavowal}: the authors decline, for their own work, a stronger, broader, more certain or more final claim, status or reading; they say how their own result, method, interpretation, claim or contribution should be read, and name as Y what it is not (for example a tested mechanism, an established finding, a general account, a definitive result, a guarantee). \emph{Approach contrast}: the use contrasts methods, designs, models, data, metrics or procedures, what was done or chosen as opposed to another approach, without qualifying how strong or general a claim is. \emph{Finding contrast}: the use states what a finding, phenomenon, behavior or property is, as opposed to something else it might have been taken for; it asserts, it does not limit a claim. \emph{Other}: none of these. Five invented examples accompanied the instructions.

\paragraph{Agreement.} The coders agree closely on whether a use is a disavowal (Table~\ref{tab:disavowal-coders}). A1's own codes, made on 80 uses sampled to include the uses on which the coders disagree, follow a broader criterion: A1 counts also uses that reject a limiting characterization of the authors' work (``our contribution is a general method rather than one calibration procedure''). Agreement with A1 is moderate whichever coder is used (Table~\ref{tab:disavowal-check}). Neither a broader written definition nor 30 of A1's codes given as examples brought the coders closer to A1 on the 50 remaining uses (Table~\ref{tab:disavowal-fewshot}), so the gap is not one of wording. A1's own codes predict A1's \emph{annoying} labels better than the coders' do, which A2's coding of the same uses will help to interpret, since A1 gave both kinds of judgment.

\begin{table}[h]
  \centering
  \footnotesize
  \setlength{\tabcolsep}{4pt}
  \begin{tabular}{lrrr}
    \hline
    & \textbf{Claude} & \textbf{GPT-6-sol} & \textbf{Both} \\
    \hline
    Uses coded as disavowals (of 947) & 147 & 148 & 129 \\
    $\kappa$, disavowal against the rest & \multicolumn{3}{c}{0.85} \\
    $\kappa$, four classes & \multicolumn{3}{c}{0.73} \\
    \hline
  \end{tabular}
  \caption{The two LLM coders on 947 uses (Claude comparison set, GPT comparison set, pool sample).}
  \label{tab:disavowal-coders}
\end{table}

\begin{table}[h]
  \centering
  \footnotesize
  \setlength{\tabcolsep}{2pt}
  \begin{tabular}{lrrrr}
    \hline
    \textbf{Codes} & \textbf{Disav.} & \textbf{$\kappa$ with A1} & \textbf{Annoying} & \textbf{AUC} \\
    \hline
    A1 & 36 & -- & 26/36 (72) & 0.83 \\
    Claude & 39 & 0.47 & 22/39 (56) & 0.70 \\
    GPT-6-sol & 41 & 0.38 & 23/41 (56) & 0.70 \\
    Both coders & 25 & 0.51 & 18/25 (72) & 0.73 \\
    Claude, broader definition & 47 & 0.53 & 26/47 (55) & 0.72 \\
    GPT-6-sol, broader definition & 52 & 0.51 & 28/52 (54) & 0.73 \\
    \hline
  \end{tabular}
  \caption{A1's coding of 80 uses compared with the coders': uses coded as disavowals, $\kappa$ with A1's codes, uses labeled \emph{annoying} by A1 among them, count/uses (\%), and the area under the ROC curve for predicting A1's \emph{annoying} labels.}
  \label{tab:disavowal-check}
\end{table}

\begin{table}[h]
  \centering
  \footnotesize
  \setlength{\tabcolsep}{4pt}
  \begin{tabular}{lrr}
    \hline
    \textbf{Instructions} & \textbf{Claude} & \textbf{GPT-6-sol} \\
    \hline
    Original & 0.56 & 0.32 \\
    Broader definition & 0.61 & 0.54 \\
    Broader, with 30 of A1's codes as examples & 0.56 & 0.58 \\
    \hline
  \end{tabular}
  \caption{$\kappa$ with A1's codes (disavowal against the rest) on the 50 of the 80 uses not used as examples.}
  \label{tab:disavowal-fewshot}
\end{table}

\paragraph{Disavowal and stance.} The two properties are nearly unrelated (Table~\ref{tab:disavowal-stance}): uses coded as disavowals have a slightly smaller stance gap, because their Y is a claim the author declines, which neither rater scores as disparaged.

\begin{table}[h]
  \centering
  \footnotesize
  \setlength{\tabcolsep}{3pt}
  \begin{tabular}{lrrrrr}
    \hline
    & \multicolumn{2}{c}{\textbf{Disavowal}} & \multicolumn{2}{c}{\textbf{Other}} & \\
    \textbf{Stance rater} & \textbf{n} & \textbf{Gap} & \textbf{n} & \textbf{Gap} & \textbf{$\rho$} \\
    \hline
    GPT-6-sol & 129 & $+$1.11 & 818 & $+$1.29 & $-$0.07 \\
    Claude agents & 81 & $+$0.86 & 468 & $+$1.21 & $-$0.09 \\
    \hline
  \end{tabular}
  \caption{Mean stance gap X$-$Y of uses coded as disavowals (both coders) and of the others, by rater, and Spearman correlation $\rho$ between disavowal and the gap.}
  \label{tab:disavowal-stance}
\end{table}

\section{Revision of the Claude and Tülu papers}
\label{app:llm-revision}

Each model revised its own first draft in response to a review, as the GPT models did (Section~\ref{sec:generation-prompts}). Table~\ref{tab:llm-revision} compares the drafts and revisions of the 100 source papers of each model, without dropping papers (Table~\ref{tab:other-models} drops those where any model's text was unusable). Sonnet 5.5 and Opus 5.5 add the postposed ``X, not Y'' in revision, whereas Haiku 4.5 does not; Haiku 4.5 and Opus 5.5 add \emph{rather than}, and Sonnet 5.5's rate of \emph{rather than} does not change. At the level of sentences, each sentence of a draft with \emph{rather than} is matched with its closest sentence in the revision. Most sentences survive with the phrase. When it is removed, in most cases no other antithesis form takes its place; when it is added, the sentence it was added to usually had none either. So the shift to ``X, not Y'' in Sonnet 5.5 and Opus 5.5 is mostly a matter of new sentences, not of rewording old ones. The Tülu 3 8B revisions use \emph{rather than} too rarely for such a comparison, and their revisions do not raise either construction.

\begin{table}[h]
  \centering
  \scriptsize
  \setlength{\tabcolsep}{2pt}
  \begin{tabular}{lrrrr}
    \hline
     & \multicolumn{2}{c}{\textbf{Change per 1k words}} & \multicolumn{2}{c}{\textbf{RT sentences}} \\
    \textbf{Model} & \textbf{RT} & \textbf{X, not Y} & \textbf{removed} & \textbf{added} \\
    \hline
    Haiku 4.5 & $+$0.16 (0.007) & $+$0.01 (0.262) & 27/399 (0) & 42/574 (3) \\
    Sonnet 5.5 & $-$0.01 (0.701) & $+$0.23 (<.001) & 39/695 (8) & 26/857 (4) \\
    Opus 5.5 & $+$0.16 (<.001) & $+$0.25 (<.001) & 92/637 (21) & 55/927 (14) \\
    \hline
  \end{tabular}
  \caption{Revision of the Claude papers: change in the mean rate from draft to revision ($p$: paired Wilcoxon test), and the sentences with \emph{rather than} (RT) removed from the draft or added in the revision, of all such sentences in the draft or the revision, with the number in which another antithesis form takes the place of the phrase (or was replaced by it).}
  \label{tab:llm-revision}
\end{table}

\section{Revision histories}
\label{app:revision}

The three datasets only show the \emph{final} text of a paper. To study whether authors themselves treat a given ``rather than'' as legitimate or as something to cut, we use the arXiv version histories of arXiv 2026 papers. Of the 1,985 candidate papers, 819 (41\%) have two or more arXiv versions, and 807 yield a usable pair of the first (v1) and the latest version. We compare the two versions of each paper and extract the diff hunks where ``rather than'' was added, removed, or kept (Appendix~\ref{app:revision}).

Revision does not remove the construction. Within the version-diff subset (793 papers), the \emph{later} arXiv version has a \emph{higher} rate than v1 (0.988 vs.\ 0.921; Table~\ref{tab:revision-test}). The mean rate rises significantly, but about as many papers go up as down, so the increase comes from a subset of papers that add many instances on revision. Diff hunks where the phrase was added outnumber those where it was removed by roughly 2:1 (1,156 vs.\ 576), and among the antithesis patterns, only ``rather than'' and, to a lesser extent, the postposed ``X, not Y'' rise on revision.

\begin{table}[t]
  \centering
  \footnotesize
  \setlength{\tabcolsep}{4pt}
  \begin{tabular}{lrr}
    \hline
     & \textbf{v1} & \textbf{Latest} \\
    \hline
    Mean rate per 1,000 words & 0.921 & 0.988 \\
    Total occurrences & 5{,}394 & 6{,}471 \\
    Median body length (words) & 6{,}747 & 7{,}572 \\
    \hline
    Paired difference & \multicolumn{2}{r}{$+$0.067} \\
    \quad 95\% CI & \multicolumn{2}{r}{0.042--0.093} \\
    \quad Wilcoxon $p$ & \multicolumn{2}{r}{$<10^{-5}$} \\
    Papers higher / lower / same & \multicolumn{2}{r}{333 / 346 / 114} \\
    \hline
  \end{tabular}
  \caption{\emph{Rather than} in the first and latest arXiv versions of the same 793 papers (body text). Paired difference: mean per-paper change, bootstrap interval, Wilcoxon signed-rank test.}
  \label{tab:revision-test}
\end{table}

Body length, references and appendices excluded: v1 800 papers, mean 7,349 and median 6,750 words; latest version 793 papers, mean 8,148 and median 7,572 words.

\paragraph{arXiv version diffs.} We use arXiv's version history: any paper posted more than once has a chance of having received an author-made revision, and a version bump often corresponds to a preprint-to-camera-ready edit for exactly the *ACL-venue papers our arXiv 2026 pre-filter selects (Table~\ref{tab:arxiv-prefilter}). Of the 1,985 arXiv 2026 candidates, 819 (41\%) have 2 or more arXiv versions. For each such paper we download the source of v1 and of the latest version, extract plain text from both, and diff them, extracting hunks where ``rather than'' was added, removed, or reworded between the two, and distinguish a \emph{surgical} removal, where the clause immediately preceding ``rather than'' survives intact, from a removal amid a larger rewrite. Collection is complete (807/819 candidates yielded a usable v1/latest pair; 12 failed on unparseable arXiv source archives). This yields 3,550 matching hunks from 807 paper pairs.

\begin{table}[t]
  \centering
  \small
  \begin{tabular}{lr}
    \hline
    \textbf{Quantity} & \textbf{Count} \\
    \hline
    Paper pairs (v1 + latest) & 807 \\
    \hline
    Mined diff hunks with phrase & 3{,}550 \\
    \quad phrase added & 1{,}156 \\
    \quad phrase kept, context reworded & 1{,}818 \\
    \quad phrase removed from hunk & 576 \\
    \quad\quad surgical removal & 86 \\
    \quad\quad removed amid larger rewrite & 490 \\
    \hline
  \end{tabular}
  \caption{arXiv version-diff mining results (807/819 candidates yielded a usable v1/latest pair).}
  \label{tab:arxiv-version-diff}
\end{table}

This holds at the level of individual diff hunks as well: hunks where the phrase was added outnumber those where it was removed by roughly 2:1 (1,156 vs.\ 576; Table~\ref{tab:arxiv-version-diff}). Papers do grow longer on revision, but the construction grows faster (Table~\ref{tab:revision-test}), so the per-word rate still increases.

\begin{figure}[t]
  \centering
  \includegraphics[width=\linewidth]{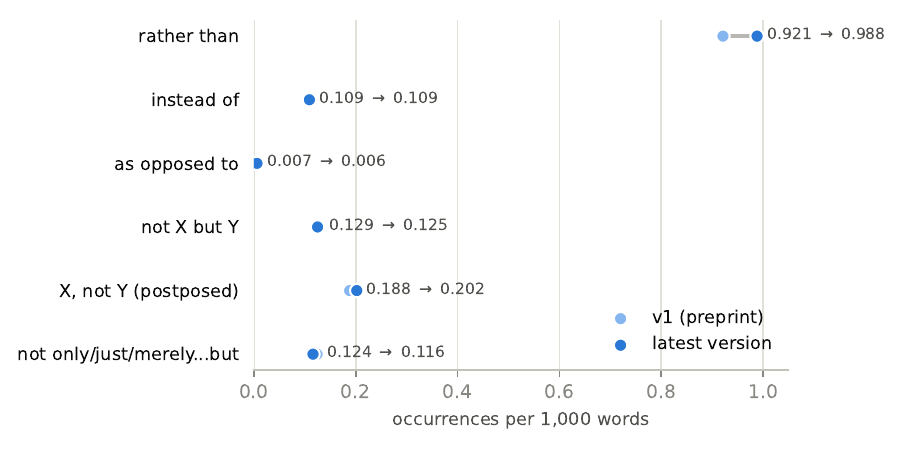}
  \caption{Occurrences per 1,000 words, first (v1) and latest arXiv versions of the same 793 papers.}
  \label{fig:antithesis-v1-latest}
\end{figure}

Figure~\ref{fig:antithesis-v1-latest} shows that the rise on revision is specific to ``rather than'' (and, to a lesser extent, the postposed ``X, not Y''): most other patterns are flat or slightly \emph{down} from v1 to the latest version, so revision does not uniformly add antithesis-family constructions -- it specifically adds more instances of the two patterns already most associated with the 2019-to-2026 rise.

Table~\ref{tab:antithesis-family-versions} gives the rate of each pattern in the first and latest versions. The revision-history mining pipelines described in Section~\ref{sec:legit-vs-annoying} currently search only for the literal phrase ``rather than''; extending them to the full construction family, and replicating the legitimate-vs-\emph{annoying} annotation for other patterns, is left for future work. 

\begin{table}[t]
  \centering
  \footnotesize
  \setlength{\tabcolsep}{3pt}
  \begin{tabular}{llrrrr}
    \hline
    \textbf{Type} & \textbf{Pattern} & \textbf{2019} & \textbf{2026} & \textbf{v1} & \textbf{latest} \\
    \hline
    Substitution & rather than & 0.134 & 0.993 & 0.921 & 0.988 \\
     & instead of & 0.165 & 0.105 & 0.109 & 0.109 \\
     & as opposed to & 0.025 & 0.007 & 0.007 & 0.006 \\
    \hline
    Negation & not X but Y & 0.097 & 0.126 & 0.129 & 0.125 \\
     & X, not Y & 0.049 & 0.201 & 0.188 & 0.202 \\
    \hline
    Emphatic & not only\ldots but & 0.079 & 0.119 & 0.124 & 0.116 \\
    \hline
  \end{tabular}
  \caption{Occurrences per 1,000 words of body text in ACL 2019, arXiv 2026, and the first (v1) and latest versions of the 793 arXiv papers with several versions.}
  \label{tab:antithesis-family-versions}
\end{table}

\paragraph{GitHub version histories.} Version histories of individual \LaTeX{} papers could show edits to single uses. We searched GitHub's code search for public repositories with an ACL-style paper (query: \verb|"\usepackage{acl}" filename:main.tex|), kept the repositories in which that file has a commit history, and mined \texttt{git log -p} for diff hunks in which \emph{rather than} was added, removed or reworded (Table~\ref{tab:github-mining}). Most repositories are single uploads with no edit history, and few have five or more commits to the paper, so the source is thin. Among the hunks that remove the phrase, we separate a \emph{surgical} removal, in which the clause before \emph{rather than} survives intact (``X rather than Y'' $\to$ ``X''), from removals within a larger rewrite of the sentence; surgical removals, the only clear evidence that an author judged the construction itself unnecessary, are a minority.

\begin{table}[t]
  \centering
  \footnotesize
  \begin{tabular}{lr}
    \hline
    \textbf{Quantity} & \textbf{Count} \\
    \hline
    Candidate repositories & 68 \\
    \quad one commit to the paper & 36 \\
    \quad five or more commits & 10 \\
    \hline
    Diff hunks with the phrase (10 repositories) & 95 \\
    \quad phrase added & 24 \\
    \quad phrase kept, context reworded & 55 \\
    \quad phrase removed & 16 \\
    \quad\quad surgical removal & 6 \\
    \quad\quad within a larger rewrite & 10 \\
    \hline
  \end{tabular}
  \caption{\emph{Rather than} in GitHub version histories of ACL-style papers.}
  \label{tab:github-mining}
\end{table}

\section{Annotation pool}
\label{app:pool}

\paragraph{Full text and body text.} The pool is drawn from full text, references and appendices included, while the rates in Section~\ref{sec:frequency} use body text only. Table~\ref{tab:annotation-strata} gives the \emph{annoying} rates without the 33 items that lie in references or appendices; the ACL 2019 vs.\ arXiv 2026 difference keeps its significance for both annotators.

\paragraph{Overlap between arXiv corpora.} arXiv 2026 uses the latest version of every paper, so for papers with several versions the arXiv 2026 and latest-version texts are identical, and a v1 text shares every sentence left unchanged in revision. An occurrence already in the pool, identified by paper, sentence and position in the sentence, is never drawn again from another corpus. The three random arXiv strata therefore share no occurrence and are combined as arXiv 2026 (all).

\paragraph{High-count papers.} Per-paper counts of \emph{rather than} are skewed: no ACL 2019 paper (of 4,863) has more than 12 occurrences, while 124 arXiv 2026 papers have 20 or more, up to 69 in one, about once every 170 words (``conditions evidence trust on claim type rather than treating all modalities as equally reliable''). The high-count stratum samples papers that lean unusually hard on the construction: 100 instances from the 22 papers with at least 35 occurrences, at most 10 per paper (1--8 in practice, mean 4.5). The threshold keeps the stratum extreme without letting one or two papers dominate it. These items are kept out of arXiv 2026 (all), since the 22 papers hold 6.5\% of the arXiv 2026 occurrences, of which a random sample of 100 would draw about seven. Table~\ref{tab:high-count} lists the papers with the most occurrences.

\begin{table}[t]
  \centering
  \footnotesize
  \begin{tabular}{lrr}
    \hline
    \textbf{Paper (arXiv id)} & \textbf{Uses} & \textbf{Per 1k words} \\
    \hline
    2608.13706 & 69 & 5.84 \\
    2608.06171 & 63 & 4.61 \\
    2606.05743 & 54 & 3.28 \\
    2608.29995 & 46 & 2.01 \\
    2607.14252 & 46 & 2.18 \\
    2604.17022 & 46 & 3.88 \\
    2605.28006 & 45 & 2.71 \\
    \hline
  \end{tabular}
  \caption{arXiv 2026 papers with the most uses of \emph{rather than}. The ACL 2019 maximum is 12.}
  \label{tab:high-count}
\end{table}

\paragraph{Batches.} The first batch was drawn in two rounds (300 and 200 items). One second-batch ACL 2019 item came from a document later excluded (Section~\ref{sec:corpora}); it is left out, and a further ACL 2019 item drawn in its place was labeled by both annotators after the rest of the pool.

\paragraph{Revision.} Uses in the first and latest arXiv versions annoy at similar rates (Table~\ref{tab:annotation-strata}). Both version strata are labeled \emph{annoying} less often than the main arXiv 2026 sample, though not significantly (Fisher's exact test, A1 $p=.13$, A2 $p=.06$), so pooling them lowers the arXiv 2026 rate slightly. 

\begin{table}[t]
  \centering
  \footnotesize
  \setlength{\tabcolsep}{3pt}
  \begin{tabular}{lrrrr}
    \hline
    \textbf{Stratum} & \textbf{N} & \textbf{A1} & \textbf{A2} & \textbf{Either} \\
    \hline
    arXiv 2026, main & 388 & 51 (13.1) & 39 (10.1) & 79 (20.5) \\
    \quad v1 & 145 & 14 (9.7) & 9 (6.2) & 22 (15.3) \\
    \quad latest & 119 & 10 (8.5) & 6 (5.0) & 15 (12.7) \\
    \hline
    \multicolumn{5}{l}{\emph{Body text only (without 33 items)}} \\
    ACL 2019 & 245 & 1.7 & 0.0 & \\
    arXiv 2026, all & 627 & 11.8 & 8.0 & \\
    High-count & 95 & 10.5 & 11.6 & \\
    \hline
  \end{tabular}
  \caption{Pool strata not shown in Table~\ref{tab:annotation}: items labeled \emph{annoying}, count (\% of the stratum); body text only: \%.}
  \label{tab:annotation-strata}
\end{table}

\section{Labels across batches}
\label{app:batches}

Table~\ref{tab:batches} recomputes the label-based results within each batch of the pool. The second batch was labeled after the first had been analyzed, so an annotator primed by the analysis would show results in the second batch that the first lacks. For A1, the first batch alone carries the corpus difference, the stance gap and the association with sentence-initial \emph{Rather than}, while the \emph{we} subject goes with annoyance only in the second batch, and the stance gap does not change between batches (logistic interaction of gap and batch, $p=.82$; for A2, $p=.54$). Across the hidden repeats, A1's labels change into \emph{annoying} three times and out of it four times, A2's twice and never.

\begin{table*}[h]
  \centering
  \footnotesize
  \begin{tabular}{lrrrr}
    \hline
     & \multicolumn{2}{c}{\textbf{A1}} & \multicolumn{2}{c}{\textbf{A2}} \\
    \textbf{Measure} & \textbf{Batch 1} & \textbf{Batch 2} & \textbf{Batch 1} & \textbf{Batch 2} \\
    \hline
    \emph{Annoying}, ACL 2019 & 0 / 143 & 4 / 100 & 0 / 144 & 0 / 99 \\
    \emph{Annoying}, arXiv 2026 & 11.5\% & 11.6\% & 5.6\% & 10.0\% \\
    \hline
    Gap X$-$Y, \emph{annoying} / legitimate & $+$2.31 / $+$0.78\rlap{$^{*}$} & $+$2.33 / $+$0.83\rlap{$^{*}$} & $+$1.79 / $+$0.91 & $+$1.43 / $+$0.90 \\
    Y, \emph{annoying} / legitimate & $-$1.03 / $-$0.31\rlap{$^{*}$} & $-$1.00 / $-$0.34\rlap{$^{*}$} & $-$0.71 / $-$0.39 & $-$0.65 / $-$0.37 \\
    \hline
    Initial \emph{Rather than} (\%) & 12 / 4\rlap{$^{*}$} & 17 / 2\rlap{$^{*}$} & 8 / 3 & 8 / 3 \\
    Subject \emph{we} (\%) & 17 / 12 & 41 / 13\rlap{$^{*}$} & 12 / 12 & 20 / 17 \\
    X adjective (\%) & 12 / 4\rlap{$^{*}$} & 7 / 4 & 15 / 5 & 22 / 2\rlap{$^{*}$} \\
    X adj.\ modifier (\%) & 0 / 0 & 4 / 1 & 0 / 0 & 10 / 0\rlap{$^{*}$} \\
    X prep.\ object (\%) & 24 / 23 & 9 / 18 & 8 / 25 & 0 / 21\rlap{$^{*}$} \\
    \emph{than} coord.\ adj. (\%) & 10 / 3 & 4 / 6 & 12 / 4 & 22 / 3\rlap{$^{*}$} \\
    Arc JJ--cc$\to$IN (\%) & 12 / 4\rlap{$^{*}$} & 4 / 6 & 12 / 5 & 22 / 3\rlap{$^{*}$} \\
    X, Y adjectives (\%) & 5 / 1 & 4 / 2 & 8 / 1 & 10 / 1\rlap{$^{*}$} \\
    Y past participle (\%) & 10 / 5 & 4 / 4 & 23 / 4\rlap{$^{*}$} & 12 / 3\rlap{$^{*}$} \\
    Y conjunct of X (\%) & 54 / 33\rlap{$^{*}$} & 35 / 37 & 77 / 32\rlap{$^{*}$} & 48 / 38 \\
    \hline
  \end{tabular}
  \caption{Label-based results within each batch of the pool. Rates: \emph{annoying} items among all items of the corpus; other rows: \emph{annoying} / legitimate arXiv 2026 items. $^{*}$\,$p<.05$ within the batch (Fisher's exact test; Mann-Whitney test, one-sided, for stance).}
  \label{tab:batches}
\end{table*}

\section{Annotator drift}
\label{app:drift}

\paragraph{Drift.} Table~\ref{tab:drift} gives the results, together with the \emph{annoying} rate across each annotator's labeling order. Most repeats keep their label (A1: 61 of 80; A2: 67 of 80), and changes into and out of \emph{annoying} are few and balanced for A1 (exact McNemar test for \emph{annoying}: A1 $p=1$, A2 $p=.5$). A1's \emph{annoying} rate is flat after the first fifth of the order. A2's rises across the order, as a jump between the two batches (5.6\% and 10.0\% of arXiv 2026 items), which were labeled on different days; within each batch it is flat. Since items appear in random order, drift shifts the overall level of \emph{annoying} labels for all strata alike, and comparisons between strata are less exposed to it than absolute rates.

\begin{table}[t]
  \centering
  \footnotesize
  \setlength{\tabcolsep}{3pt}
  \begin{tabular}{lrr}
    \hline
     & \textbf{A1} & \textbf{A2} \\
    \hline
    \multicolumn{3}{l}{\textit{Hidden repeats (80 items)}} \\
    Same label on both passes & 61 (76\%) & 67 (84\%) \\
    \texttt{legitimate} $\to$ \texttt{unsure} & 7 & 3 \\
    \texttt{unsure} $\to$ \texttt{legitimate} & 4 & 8 \\
    other $\to$ \emph{annoying} & 3 & 2 \\
    \emph{annoying} $\to$ other & 4 & 0 \\
    \hline
    \multicolumn{3}{l}{\textit{Labeling order (garbled excluded)}} \\
    \emph{Annoying} rate by fifth (\%) & 6, 11, 10, 12, 8 & 5, 6, 8, 6, 9 \\
    Odds ratio, first to last & 0.9 ($p=.88$) & 2.5 ($p=.07$) \\
    \hline
  \end{tabular}
  \caption{Annotator drift. \emph{Repeats}: early items re-labeled blind later. \emph{Order}: \emph{annoying} rate in successive fifths of each annotator's items, and a logistic regression on position with stratum fixed effects.}
  \label{tab:drift}
\end{table}

\section{A5's labels}
\label{app:a5}

A5 labeled 51 items of the annotation pool, drawn from all strata. Table~\ref{tab:a5} compares these labels with A1's and A2's on the same items. A5 labeled \emph{annoying} 12 of 16 ACL 2019 items and 17 of 26 random arXiv 2026 items, about the same share in both corpora (Fisher's exact test, $p=.73$).

\begin{table}[t]
  \centering
  \footnotesize
  \setlength{\tabcolsep}{3pt}
  \begin{tabular}{lrrr}
    \hline
     & \textbf{A1--A2} & \textbf{A1--A5} & \textbf{A2--A5} \\
    \hline
    Raw agreement & 76\% & 26\% & 20\% \\
    Cohen's $\kappa$ & 0.15 & $-$0.00 & $-$0.03 \\
    \emph{Annoying}: 1st / 2nd / both & 4 / 1 / 0 & 4 / 35 / 3 & 1 / 35 / 0 \\
    \hline
  \end{tabular}
  \caption{Agreement on the 50 items A5 labeled that no annotator marked \texttt{garbled} (raw agreement and $\kappa$ over three labels).}
  \label{tab:a5}
\end{table}

\section{Further correlates of annoyance}
\label{app:other-factors}

The analyses below test impressions from reviewing the pool, on the arXiv 2026 items of both batches. Items marked \texttt{garbled} are left out.

\paragraph{Research domain.} Reviewing the pool suggested that uses in papers on a scientific domain outside core NLP read as more legitimate. We tested this with a procedure fixed before the domain labels existed. GPT-6-sol labeled each arXiv 2026 paper behind the pool, from its title and abstract, as core NLP (models, methods, evaluation, LLM behavior), applied to another field (e.g.\ medicine, law, education), or linguistics. We compared the \emph{annoying} rate in core NLP papers with that in the other two groups combined, with a logistic regression controlling for stratum. The predicted difference does not appear (Table~\ref{tab:domain}). Uses in papers from other domains are labeled \emph{annoying} more often, significantly so for A2 only; since this direction was not predicted, we treat it as a hypothesis for new labels. 

\begin{table}[t]
  \centering
  \footnotesize
  \setlength{\tabcolsep}{4pt}
  \begin{tabular}{lrrrr}
    \hline
     & \textbf{Core NLP} & \textbf{Other} & \textbf{OR} & $p$ \\
    \hline
    A1 & 69/603 (11\%) & 18/132 (14\%) & 1.2 & .46 \\
    A2 & 46/602 (8\%) & 19/132 (14\%) & 2.1 & .018 \\
    Either & 102/601 (17\%) & 31/132 (23\%) & 1.5 & .082 \\
    \hline
  \end{tabular}
  \caption{\emph{Annoying} arXiv 2026 items by the domain of their paper (403 core NLP, 70 applied, 8 linguistics papers). OR: other vs.\ core NLP, logistic regression with stratum; $p$: Fisher's exact test. Predicted direction: one-sided $p\geq.77$.}
  \label{tab:domain}
\end{table}

\paragraph{Density and other AI tells.} Two further impressions are that a use annoys more when the construction recurs nearby, and when the surrounding text shows other marks of LLM writing. Table~\ref{tab:tells} tests both on the text the annotators saw (the sentence and its neighbors). For AI tells we used published definitions: the style words among the excess vocabulary of \citet{kobak2025delving}, and the named tell sets of \citet{druck-paredes-smith-2026-ai-tells} (marketing, hype, intensifier, precision, evaluative, transition, tradeoff and importance terms; their corrective-framing set overlaps the antithesis family and is left out). Neither is associated with annoyance for either annotator; sentence-initial emphatic adverbs (\emph{Crucially,}) and em dashes are too rare in the extracted text to test. Another \emph{rather than} or another antithesis pattern nearby does not raise the \emph{annoying} rate either. At the level of the whole paper, which annotators never saw, uses from papers with a higher \emph{rather than} rate are \emph{annoying} more often for A2 and not for A1; this comparison was not planned.

\begin{table}[t]
  \centering
  \footnotesize
  \setlength{\tabcolsep}{3pt}
  \begin{tabular}{lrrrr}
    \hline
    \textbf{Shown text} & \textbf{A1} & $p$ & \textbf{A2} & $p$ \\
    \hline
    Another \emph{rather than} & 9 / 12 & .41 & 7 / 9 & .57 \\
    Other antithesis & 17 / 12 & .32 & 10 / 9 & .78 \\
    Graphite tell & 11 / 12 & 1.0 & 9 / 9 & .85 \\
    Kobak words /100 & 5.1 / 4.8 & .10 & 4.4 / 4.8 & .80 \\
    \hline
    \multicolumn{5}{l}{\textit{Whole paper}} \\
    \emph{Rather than} /1k & 1.48 / 1.56 & .46 & 1.61 / 1.53 & .013 \\
    \hline
  \end{tabular}
  \caption{Density and AI tells, arXiv 2026 items. First three rows: \emph{annoying} rate (\%) with / without the feature, Fisher's exact test. Kobak: mean rate, \emph{annoying} / other items, one-sided logistic test (\citealp{kobak2025delving}; Graphite: \citealp{druck-paredes-smith-2026-ai-tells}). Whole paper: median rate, \emph{annoying} / other items, logistic regression on $\log(1+\mathrm{rate})$ with stratum.}
  \label{tab:tells}
\end{table}

\paragraph{Clustering by paper.} Annotators saw items one at a time, in random order, without the source paper. Still, A2's \emph{annoying} labels concentrate in some papers: among papers with two or more items, pairs of \emph{annoying} items from the same paper are far more frequent than under shuffled labels (Table~\ref{tab:cluster}). A1's are not. We also asked whether a straw man elsewhere in a paper goes with A2's annoyance at the paper's other uses, a pairing in which A2 never saw the straw man. It does at first sight, but the association disappears in a logistic regression that includes A1's label on the same item and the paper's \emph{rather than} rate. A2's annoyance across a paper thus goes with how heavily the paper uses the construction (Table~\ref{tab:tells}).

\begin{table}[t]
  \centering
  \footnotesize
  \setlength{\tabcolsep}{3pt}
  \begin{tabular}{lrrr}
    \hline
     & \textbf{Papers} & \textbf{Same-paper pairs} & $p$ \\
    \hline
    A1 & 129 & 11 vs.\ 8.6 & .25 \\
    A2 & 129 & 24 vs.\ 6.4 & $<$.001 \\
    \hline
    \multicolumn{4}{l}{\textit{A2 \emph{annoying}, by A1's codes elsewhere in the paper}} \\
    \multicolumn{2}{l}{\quad straw man / none} & 12/64 vs.\ 8/105 & .047 \\
    \multicolumn{2}{l}{\quad controlled} & OR 1.5 & .51 \\
    \hline
  \end{tabular}
  \caption{Clustering of \emph{annoying} labels by paper, arXiv 2026 papers with two or more items. \emph{Same-paper pairs}: observed vs.\ expected under 5,000 shuffles of labels within strata. Last rows: Fisher's exact test, then a logistic regression with A1's label on the item, the paper's \emph{rather than} rate and the high-count stratum, paper-clustered standard errors.}
  \label{tab:cluster}
\end{table}

\paragraph{Lexical correlates.} Every token and 2--3-gram, in the sentence and separately in its context, is scored by the log-odds ratio with an informative prior \citep{monroe2008fightin}, comparing the share of \emph{annoying} and legitimate items that contain it, with the prior taken from all arXiv 2026 instances. No term survives Benjamini-Hochberg correction over all terms in any label set. The highest-scoring terms (Table~\ref{tab:lexical}) are thus candidates to test on new labels. A1's \emph{annoying} items more often contain \emph{intended} (purpose framing: ``our approach is intended to support, rather than replace, educators''), \emph{rather than relying}, one of the LLM frames of \citet{druck-paredes-smith-2026-ai-tells}, and \emph{we}, the first-person subject of the syntactic scan (Section~\ref{sec:correlates}). A2's more often contain \emph{so}, mostly in ``[evidence], so [claim] is X rather than Y''. A definite noun phrase after the construction (\emph{rather than the}), which names a specific alternative, is rare in \emph{annoying} items for both annotators.

\begin{table}[t]
  \centering
  \footnotesize
  \setlength{\tabcolsep}{4pt}
  \begin{tabular}{lr@{\,/\,}lr@{\,/\,}lr@{\,/\,}l@{\hspace{9pt}}}
    \hline
    \textbf{Term} & \multicolumn{2}{c}{\textbf{A1}} & \multicolumn{2}{c}{\textbf{A2}} & \multicolumn{2}{c}{\textbf{Either}} \\
    \hline
    Items & 87 & 598 & 66 & 582 & 134 & 486 \\
    \hline
    \emph{intended} & 8 & 1 & 0 & 1 & 5 & 1 \\
    \emph{rather than relying} & 8 & 2 & 2 & 2 & 5 & 2 \\
    \emph{we} & 34 & 18 & 21 & 21 & 29 & 19 \\
    \emph{so} & 10 & 8 & 17 & 7 & 11 & 7 \\
    \emph{rather than the} & 3 & 5 & 0 & 5 & 2 & 5 \\
    \emph{not} (context) & 23 & 15 & 24 & 16 & 22 & 14 \\
    \hline
  \end{tabular}
  \caption{Candidate terms: \% of \emph{annoying} / legitimate arXiv 2026 items containing each term, in the sentence unless marked. No term has $q<.05$ in the scan (Benjamini-Hochberg over 3,475--3,948 terms per label set).}
  \label{tab:lexical}
\end{table}

\paragraph{Blatant straw men.} The six items both coders coded as straw men include the clearest cases in the pool, where the rejected alternative is a lapse no author would admit to: ``the choice is load-bearing and is disclosed rather than defended'', ``this is counted rather than assumed'', ``reflects clarity rather than circular validation''. Both annotators labeled these three \emph{annoying}. Reviewing the pool suggested that a few such cases may shape how readers perceive the construction as a whole. We made three attempts to test this without further annotation. First, we scored every item's Y by its maximum embedding similarity to the agreed straw men and to the invented examples of the straw-man definition, and measured how well the score ranks straw men and \emph{annoying} items (Table~\ref{tab:blatant}). It ranks A1's straw men and A1's \emph{annoying} items above chance, and A2's only when the whole clause ``X rather than Y'' is scored. The top-ranked items mostly share topic words with the seeds (\emph{answers}, \emph{random}, \emph{diagnostic}), and few of them reject a lapse, so similarity does not isolate the blatant cases. Second, since items were shown in random order, we tested whether meeting one of the six raised the \emph{annoying} rate of the next five items: it did not, for either annotator, although with six such items the test has little power (Table~\ref{tab:blatant}). Third, the blatant cases share a surface form, a past participle on both sides (``X-ed rather than Y-ed''), which a part-of-speech tagger finds without a word list. The form covers ordinary contrasts too (``spoken rather than written''), and the blatant cases are a minority of its matches; its share of uses still rises from ACL 2019 to arXiv 2026, and further in GPT papers, most of all after revision (Table~\ref{tab:participle}). 

\begin{table}[t]
  \centering
  \footnotesize
  \setlength{\tabcolsep}{4pt}
  \begin{tabular}{lrrr}
    \hline
     & \textbf{A1} & \textbf{A2} & \textbf{Either} \\
    \hline
    \multicolumn{4}{l}{\textit{Similarity to blatant straw men, AUC}} \\
    \quad Y, straw-man codes & .67\rlap{$^{*}$} & & \\
    \quad Y, \emph{annoying} & .66\rlap{$^{*}$} & .53 & .61\rlap{$^{*}$} \\
    \quad clause, \emph{annoying} & .64\rlap{$^{*}$} & .61\rlap{$^{*}$} & .63\rlap{$^{*}$} \\
    \hline
    \multicolumn{4}{l}{\textit{\emph{Annoying} in the next five items, \%}} \\
    \quad after / otherwise & 9 / 11 & 5 / 9 & \\
    \hline
  \end{tabular}
  \caption{Blatant straw men. AUC: how well the maximum embedding similarity (\texttt{text-embedding-3-large}) to the six agreed straw men (and, for Y, the definition's examples) ranks A1's straw-man codes or \emph{annoying} arXiv 2026 items; $^{*}$\,$p<.05$, permutation test. Next five items: \emph{annoying} rate after meeting one of the six vs.\ otherwise (Fisher's exact test, $p=1$ for both).}
  \label{tab:blatant}
\end{table}

\begin{table}[t]
  \centering
  \footnotesize
  \setlength{\tabcolsep}{4pt}
  \begin{tabular}{lrrr}
    \hline
    \textbf{Source} & \textbf{\emph{Rather than}} & \textbf{X-ed/Y-ed} & \textbf{Share} \\
    \hline
    ACL 2019 & 3{,}087 & 4 & 0.13\% \\
    arXiv 2026 & 14{,}539 & 82 & 0.56\% \\
    GPT-4o & 104 & 1 & 0.96\% \\
    \quad revised & 73 & 1 & 1.4\% \\
    GPT-5.6-sol & 5{,}878 & 74 & 1.3\% \\
    \quad revised & 7{,}636 & 115 & 1.5\% \\
    GPT-6-sol & 5{,}007 & 118 & 2.4\% \\
    \quad revised & 4{,}390 & 160 & 3.6\% \\
    \hline
  \end{tabular}
  \caption{``X-ed rather than Y-ed'': uses of \emph{rather than} with a past participle (spaCy tag VBN) right before \emph{rather} and after \emph{than}. Full text; GPT rows: all 400 papers per model.}
  \label{tab:participle}
\end{table}

\section{Correlates of annoyance in the GPT comparison set}
\label{app:batch3}

Table~\ref{tab:batch3-correlates} tests the syntactic patterns of Table~\ref{tab:syntax} and the stance comparisons on the GPT comparison set, with arXiv 2026 and GPT items pooled and corpus controlled (Cochran-Mantel-Haenszel tests for patterns; for stance, Mann-Whitney tests, one-sided as in the pool). Sentence-initial \emph{Rather than} is too rare in this set to test. A scan of all patterns within the set finds, at $q<.05$, patterns of a present-tense main clause with a \emph{we} subject for A1, and of passive clauses and past participles for A2. Table~\ref{tab:batch3-cohesion} repeats the cohesion test separately for arXiv 2026 and GPT items, since nearest neighbors would otherwise reflect the source; for GPT items, papers written from the same source paper count as the same paper.

\begin{table}[h]
  \centering
  \footnotesize
  \setlength{\tabcolsep}{2.5pt}
  \begin{tabular}{lr@{\,/\,}lr@{\,/\,}lr@{\,/\,}l@{\hspace{9pt}}}
    \hline
    \textbf{Feature} & \multicolumn{2}{c}{\textbf{A1}} & \multicolumn{2}{c}{\textbf{A2}} & \multicolumn{2}{c}{\textbf{Either}} \\
    \hline
    Items & 56 & 251 & 27 & 293 & 72 & 221 \\
    \hline
    Subject \emph{we} & 29 & 9\rlap{$^{*}$} & 11 & 12 & 22 & 10\rlap{$^{*}$} \\
    X adjective & 12 & 4 & 11 & 3 & 11 & 2\rlap{$^{*}$} \\
    X adj.\ modifier & 2 & 0 & 0 & 0 & 1 & 0 \\
    X prep.\ object & 27 & 21 & 15 & 24 & 22 & 22 \\
    \emph{than} coord.\ adj. & 12 & 3\rlap{$^{*}$} & 7 & 3 & 10 & 2\rlap{$^{*}$} \\
    Arc JJ--cc$\to$IN & 12 & 4\rlap{$^{*}$} & 7 & 3 & 10 & 2\rlap{$^{*}$} \\
    X, Y adjectives & 4 & 3 & 4 & 1 & 3 & 1 \\
    Y past participle & 9 & 5 & 19 & 4\rlap{$^{*}$} & 12 & 3\rlap{$^{*}$} \\
    Y conjunct of X & 54 & 40 & 67 & 40\rlap{$^{*}$} & 57 & 38\rlap{$^{*}$} \\
    \hline
    Gap X$-$Y & 1.55 & 1.04\rlap{$^{*}$} & 1.48 & 1.13 & 1.51 & 0.95\rlap{$^{*}$} \\
    Y rating & $-$.64 & $-$.48 & $-$.74 & $-$.51 & $-$.65 & $-$.44\rlap{$^{*}$} \\
    \hline
  \end{tabular}
  \caption{GPT comparison set, arXiv 2026 and GPT items: \emph{annoying} / legitimate, \% of items (gap and Y: mean stance rating). $^{*}$\,$p<0.05$.}
  \label{tab:batch3-correlates}
\end{table}

\begin{table}[h]
  \centering
  \footnotesize
  \setlength{\tabcolsep}{3pt}
  \begin{tabular}{llrrr@{\hspace{9pt}}}
    \hline
    \textbf{Items} & \textbf{Unit} & \textbf{A1} & \textbf{A2} & \textbf{Either} \\
    \hline
    arXiv 2026 & Items & 32 / 121 & 9 / 153 & 36 / 111 \\
     & Chance & 20\% & 5\% & 24\% \\
     & X & 31\%\rlap{$^{*}$} & 9\% & 34\%\rlap{$^{*}$} \\
     & Y & 41\%\rlap{$^{*}$} & 24\%\rlap{$^{*}$} & 42\%\rlap{$^{*}$} \\
     & Sentence & 41\%\rlap{$^{*}$} & 0\% & 44\%\rlap{$^{*}$} \\
     & Paragraph & 32\%\rlap{$^{*}$} & 4\% & 33\% \\
    \hline
    GPT & Items & 24 / 130 & 18 / 140 & 36 / 110 \\
     & Chance & 15\% & 11\% & 24\% \\
     & X & 22\% & 31\%\rlap{$^{*}$} & 42\%\rlap{$^{*}$} \\
     & Y & 33\%\rlap{$^{*}$} & 28\%\rlap{$^{*}$} & 43\%\rlap{$^{*}$} \\
     & Sentence & 41\%\rlap{$^{*}$} & 27\%\rlap{$^{*}$} & 48\%\rlap{$^{*}$} \\
     & Paragraph & 19\% & 21\%\rlap{$^{*}$} & 32\% \\
    \hline
  \end{tabular}
  \caption{GPT comparison set: share of \emph{annoying} items' five nearest neighbors that are \emph{annoying}, as in Table~\ref{tab:cohesion}. $^{*}$\,$p<0.05$ (label permutations).}
  \label{tab:batch3-cohesion}
\end{table}

\section{Perceived LLM authorship}
\label{app:authorship}

A1 and A2 judged 142 passages: 107 from just before or after the context they had labeled and 35 containing it. Table~\ref{tab:authorship-detail} relates their hunches to the annoyance labels at the level of papers and of single passages. Among arXiv 2026 passages, those containing \emph{rather than} read as LLM-written to both annotators about twice as often as those without it, also among passages from outside the labeled context. Since both annotators knew the hypotheses, this result could reflect their expectation; the judgments of A3 and A5, who did not know them, show the same effect (Table~\ref{tab:authorship-pool}).

\begin{table}[h]
  \centering
  \footnotesize
  \setlength{\tabcolsep}{3pt}
  \begin{tabular}{llrr}
    \hline
    \textbf{Hunches} & \textbf{Labels} & \textbf{Papers} & \textbf{Passages} \\
    \hline
    A1 & A1 & $-$0.01 & $-$0.06 \\
    A1 & A2 & \textbf{0.21} & \textbf{0.26} \\
    A2 & A2 & \textbf{0.18} & 0.16 \\
    A2 & A1 & 0.02 & 0.01 \\
    \hline
    \textbf{arXiv 2026} & \textbf{n} & \textbf{A1} & \textbf{A2} \\
    \hline
    with \emph{rather than} & 52 & \textbf{73\%} & \textbf{50\%} \\
    without & 76 & 34\% & 17\% \\
    \hline
  \end{tabular}
  \caption{Authorship hunches and annoyance. Upper part, Spearman $\rho$: \emph{Papers}, mean hunch vs.\ share of a paper's pool items labeled \emph{annoying} (125 papers); \emph{Passages}, hunch vs.\ the label of the item next to which a passage was drawn (107 passages). Lower part: passages rated LLM-leaning. Bold: $p<.05$.}
  \label{tab:authorship-detail}
\end{table}

\section{Checks of the stance ratings}
\label{app:stance-check}

A1 judged 80 arXiv 2026 items, shuffled, with X and Y marked and the surrounding context shown, and answered which alternative the author presents more favorably: X, Y, or neither (both presented as plain options, without evaluation, or equally). Forty items were drawn at random from the 152 that GPT-6-sol rated inverted (X below Y) and forty from the 596 it did not; the ratings and annoyance labels were not shown. Table~\ref{tab:stance-check} gives the answers. Agreement between the inverted rating and a judgment that the author favors Y is 78\% ($\kappa=0.55$).

\begin{table}[h]
  \centering
  \footnotesize
  \begin{tabular}{lrrr}
    \hline
    \textbf{GPT-6-sol} & \textbf{Favors X} & \textbf{Favors Y} & \textbf{Neither} \\
    \hline
    Inverted & 1 & 24 & 15 \\
    Not inverted & 20 & 2 & 18 \\
    \hline
  \end{tabular}
  \caption{A1's judgment of which alternative the author favors, for 40 items GPT-6-sol rated inverted and 40 it did not.}
  \label{tab:stance-check}
\end{table}

\paragraph{A second rater.} To see whether the ratings depend on the rating model, Claude (Sonnet 5.5, run as independent agents) rated 549 uses again, with the same instructions, passages and X and Y spans, and without the GPT-6-sol ratings or any label: the 310 uses of the Claude comparison set, and a sample of the pool (120 uses that either annotator labeled \emph{annoying} and 119 that neither did). The two raters agree well (Table~\ref{tab:stance-raters}): they differ by at most one point on nearly every use, order the uses alike, and agree on which uses are inverted more closely than GPT-6-sol agrees with A1 above. Their levels differ little, except that the second rater finds the GPT-6-sol uses less disparaging than GPT-6-sol itself does (Table~\ref{tab:stance-sources}); it nevertheless orders the sources as GPT-6-sol does, and it rates the uses of Claude's own first drafts as GPT-6-sol does, so it shows no sign of favoring text from its own family. The association of stance with annoyance, the main use of the ratings, replicates in the pool sample with the second rater, and with the mean of both raters (Table~\ref{tab:stance-annoy}). In the Claude comparison set the association is weak under either rater and under their mean, so that set cannot show whether stance accounts for the annoyance of Claude's uses.

\begin{table}[h]
  \centering
  \footnotesize
  \setlength{\tabcolsep}{4pt}
  \begin{tabular}{lrrrr}
    \hline
    & \textbf{$\kappa_w$} & \textbf{Spearman} & \textbf{Within 1} & \textbf{Mean diff.} \\
    \hline
    X (affirmed) & 0.77 & 0.76 & 98\% & $-$0.12 \\
    Y (rejected) & 0.72 & 0.73 & 97\% & $+$0.02 \\
    Gap X$-$Y & 0.81 & 0.80 & 81\% & $-$0.14 \\
    Inverted (X$<$Y), $\kappa$ & 0.71 & & & \\
    \hline
  \end{tabular}
  \caption{Agreement of the Claude agents with GPT-6-sol on 549 uses: quadratic weighted $\kappa$ ($\kappa_w$), Spearman correlation, share of uses within one point, and mean difference (Claude agents minus GPT-6-sol).}
  \label{tab:stance-raters}
\end{table}

\begin{table}[h]
  \centering
  \footnotesize
  \setlength{\tabcolsep}{4pt}
  \begin{tabular}{lrrr}
    \hline
    & \textbf{Uses} & \textbf{GPT-6-sol} & \textbf{Claude} \\
    \hline
    ACL 2019 originals & 50 & $+$1.22 & $+$1.04 \\
    Claude, first draft & 85 & $+$0.81 & $+$0.86 \\
    Claude, revised & 75 & $+$0.81 & $+$0.61 \\
    GPT-6-sol, first draft & 50 & $+$1.88 & $+$1.52 \\
    GPT-6-sol, revised & 50 & $+$1.84 & $+$1.54 \\
    \hline
  \end{tabular}
  \caption{Mean stance gap X$-$Y of the uses in the Claude comparison set, by rater.}
  \label{tab:stance-sources}
\end{table}

\begin{table}[h]
  \centering
  \footnotesize
  \setlength{\tabcolsep}{2pt}
  \begin{tabular}{llrrr}
    \hline
    \textbf{Set} & \textbf{Rater} & \textbf{Gap: ann. / other} & \textbf{AUC} & \textbf{$p$} \\
    \hline
    Pool sample & GPT-6-sol & $+$1.99 / $+$0.81 & 0.65 & $<$.001 \\
     & Claude & $+$1.79 / $+$0.81 & 0.65 & $<$.001 \\
     & mean & $+$1.89 / $+$0.81 & 0.66 & $<$.001 \\
    Claude set (A1) & GPT-6-sol & $+$1.42 / $+$1.18 & 0.53 & .46 \\
     & Claude & $+$1.40 / $+$0.98 & 0.58 & .08 \\
     & mean & $+$1.41 / $+$1.08 & 0.55 & .24 \\
    \hline
  \end{tabular}
  \caption{Stance gap X$-$Y of \emph{annoying} (ann.) and other uses, by rater (Claude: the agents). Pool sample: 120 uses either annotator labeled \emph{annoying} and 119 neither did. Claude comparison set: A1's 48 \emph{annoying} and 262 other uses. AUC: probability that an \emph{annoying} use has a larger gap than another use (Mann-Whitney test).}
  \label{tab:stance-annoy}
\end{table}

\section{Straw-man coding: agreement and LLM coders}
\label{app:strawman}

Coding straw men is hard. The four human coders apply very different thresholds, from 7 to 42 straw men among the same 151 items, and agree little on which items are straw men (Table~\ref{tab:strawman-kappa}). Most of what they share is a few clear cases.

\begin{table}[h]
  \centering
  \footnotesize
  \begin{tabular}{lrrrr}
    \hline
     & \textbf{A1} & \textbf{A2} & \textbf{A3} & \textbf{A4} \\
    \hline
    Straw men (of 151) & 42 & 7 & 40 & 12 \\
    \hline
    $\kappa$ with A2 & 0.18 & & & \\
    $\kappa$ with A3 & 0.16 & 0.10 & & \\
    $\kappa$ with A4 & 0.11 & 0.27 & 0.12 & \\
    \hline
  \end{tabular}
  \caption{Straw-man coders: number of items each codes straw man, and Cohen's $\kappa$ between coders (straw man vs.\ other, all 151 items).}
  \label{tab:strawman-kappa}
\end{table}

LLM coders do no better. GPT-6 models agree with the human coders about as well as the coders agree with each other: $\kappa=0.14$--$0.17$ with A1 given the definition alone, and 0.27 with A1 and 0.22 with A2 given human-coded examples from other folds (five-fold cross-validation).

\section{Exposure effects in annotation}
\label{app:exposure}

Annotating the pool made two exposure effects apparent to A1. First, the construction as a whole grew more annoying with repeated encounters. Repetition is often associated with increased liking \citep{zajonc1968attitudinal}, but this effect weakens and reverses with many exposures \citep{bornstein1989exposure}, and responses that strengthen with repetition are described as sensitization \citep{groves1970habituation}. Second, A1 came to recognize specific subpatterns, such as contrasts between abstract near-antonyms, that had not stood out at first. This matches accounts of perceptual learning as increasing attention to diagnostic features \citep{goldstone1998perceptual}; once noticed, such a pattern can also seem more frequent than it is, a tendency \citet{zwicky:2007:why,zwicky2005just} calls the frequency illusion. A further mechanism may widen the category itself: when members of a category become rare, people extend the category to cases they would previously have excluded \citep{levari2018prevalence}, a shift that range-frequency accounts of category judgment help explain \citep{parducci1965category}. Since \emph{annoying} items make up a small share of the pool, this predicts a gradual widening of an annotator's \emph{annoying} category, and \citet{levari2018prevalence} found that warnings and incentives did not prevent such shifts. We therefore measure drift directly with repeated items and labeling order (Table~\ref{tab:drift}). A1's overall rate shows no net shift across the labeling order, although A1's own impression was of growing annoyance, and A1's ACL 2019 \emph{annoying} labels all come from the second batch; A2's \emph{annoying} rate is higher in the second batch, labeled on another day, than in the first, while the second batch has a larger share of arXiv 2026 items, which the prevalence account predicts would lower it. Annotators may thus differ in their susceptibility to these effects.

These effects also bear on what the annotation measures. Readers of recent NLP papers encounter the construction often, so a reader sensitized by repeated exposure is a realistic reader, and so is a first-time reader. Planned annotators will begin with short sessions, which gives a sample of judgments made before heavy exposure, and varying the number of instances an annotator sees before the scored items would test the exposure effect directly. 

\paragraph{Exposure in the Claude comparison set.} Uses in revised papers could annoy more simply because a revised text repeats what the reader has just seen. The data argue against this (Table~\ref{tab:exposure}). Near-identical sentences from a draft and its revision are rare in the set, and none was labeled \emph{annoying}. A1's \emph{annoying} rate does not rise across the labeling order (about 17\% in its first half and 18\% in its second). And for Claude, the higher rate of revised uses remains, with a similar size, when the position in the session and the number of earlier uses from the same paper are controlled; neither of these two predicts annoyance. GPT-6-sol's revisions, which A1 also saw after its drafts, do not annoy more.

\begin{table}[h]
  \centering
  \footnotesize
  \setlength{\tabcolsep}{2pt}
  \begin{tabular}{lrrrr}
    \hline
    & \textbf{Revised} & \textbf{Ctrl.} & \textbf{Position} & \textbf{Earlier} \\
    \hline
    Claude & $+$0.94 (0.03) & $+$0.90 (0.05) & $+$0.35 (0.63) & $-$0.12 (0.26) \\
    GPT-6-sol & $-$0.51 (0.32) & $-$0.35 (0.50) & $-$1.86 (0.15) & $-$0.13 (0.45) \\
    \hline
  \end{tabular}
  \caption{Logistic regression of A1's \emph{annoying} label on revised versus first draft, alone (\emph{Revised}) and controlled (\emph{Ctrl.}) for the position of the item in the labeling order (\emph{Position}) and the number of earlier-labeled uses from the same source paper (\emph{Earlier}): coefficients (log-odds) and $p$ values; the last two columns are the coefficients of the controls. Of 97 pairs of draft and revised uses from the same paper and model, 5 are near-identical (similarity at least 0.6), none \emph{annoying}.}
  \label{tab:exposure}
\end{table}

\end{document}